\pdfoutput=1
\documentclass[11pt]{article}

\usepackage[final]{acl}

\usepackage{times}
\usepackage{latexsym}
\usepackage[T1]{fontenc}
\usepackage[utf8]{inputenc}

\usepackage{microtype}
\usepackage{graphicx}
\usepackage{url}
\usepackage{booktabs}
\usepackage{amsfonts}
\usepackage{nicefrac}
\usepackage{listings}
\usepackage{amsmath}
\usepackage{tabularx}
\usepackage{multirow}
\usepackage[table]{xcolor}
\usepackage{array}
\usepackage{ragged2e}
\usepackage{makecell}
\usepackage{amssymb}
\usepackage[most]{tcolorbox}
\usepackage{adjustbox}

\newcolumntype{L}[1]{>{\RaggedRight\arraybackslash}p{#1}}
\newcolumntype{Y}{>{\RaggedRight\arraybackslash}X}
\newcolumntype{C}[1]{>{\centering\arraybackslash}p{#1}}

\definecolor{tablehead}{RGB}{235,242,250}
\definecolor{groupfill}{RGB}{246,248,251}
\definecolor{bestfill}{RGB}{226,239,218}

\newcommand{\best}[1]{\textbf{#1}}

\newcommand{\no}{$\times$}
\newcommand{\yes}{\checkmark}

\tcbuselibrary{listings,breakable,skins}

\definecolor{promptbg}{RGB}{248,250,252}
\definecolor{promptborder}{RGB}{203,213,225}

\definecolor{systemblue}{RGB}{37,99,235}
\definecolor{usergreen}{RGB}{22,163,74}
\definecolor{judgepurple}{RGB}{124,58,237}
\definecolor{ruleamber}{RGB}{217,119,6}

\definecolor{systembg}{RGB}{239,246,255}
\definecolor{userbg}{RGB}{240,253,244}
\definecolor{judgebg}{RGB}{245,243,255}
\definecolor{rulebg}{RGB}{255,251,235}

\newtcblisting{promptbox}[2]{
  enhanced,
  breakable,
  listing only,
  listing options={
    basicstyle=\ttfamily\scriptsize,
    breaklines=true,
    breakatwhitespace=false,
    columns=fullflexible,
    keepspaces=true,
    showstringspaces=false,
    tabsize=2
  },
  title={#1},
  colback=#2!6,
  colframe=#2!70!black,
  coltitle=white,
  fonttitle=\bfseries\scriptsize,
  boxed title style={
    colback=#2!75!black,
    sharp corners
  },
  left=5pt,
  right=5pt,
  top=5pt,
  bottom=5pt,
  boxrule=0.45pt,
  arc=2pt,
  borderline west={1.4pt}{0pt}{#2!85!black},
  before skip=0.65em,
  after skip=0.65em
}

\definecolor{twitterblue}{RGB}{29,155,240}
\definecolor{twitterlight}{RGB}{232,245,253}

\newtcolorbox{postbox}[2][]{
  enhanced,
  breakable,
  colback=twitterlight,
  colframe=twitterblue!60!black,
  colbacktitle=twitterblue!80!black,
  coltitle=white,
  fonttitle=\bfseries\footnotesize,
  title={\parbox{\dimexpr\linewidth-12pt\relax}{#2}},
  toptitle=4pt,
  bottomtitle=4pt,
  lefttitle=6pt,
  righttitle=6pt,
  left=7pt,
  right=7pt,
  top=7pt,
  bottom=7pt,
  boxrule=0.4pt,
  arc=2pt,
  borderline west={1.2pt}{0pt}{twitterblue!80!black},
  before skip=0.65em,
  after skip=0.65em,
  #1
}

\newcommand{\postentry}[5]{%
\begin{tcolorbox}[
  enhanced,
  colback=white,
  colframe=twitterblue!22,
  boxrule=0.25pt,
  arc=2pt,
  left=6pt,
  right=6pt,
  top=6pt,
  bottom=6pt,
  before skip=6pt,
  after skip=6pt
]
\begin{tabularx}{\linewidth}{
  @{}
  >{\RaggedRight\arraybackslash}X
  @{\hspace{10pt}}
  >{\centering\arraybackslash}p{0.36\linewidth}
  @{}
}
\vspace{0pt}
{\bfseries #1}\quad{\scriptsize\color{black!60}#2}\par
\vspace{3pt}
#3
&

\adjustbox{valign=t}{%
  \includegraphics[width=\linewidth,height=#5,keepaspectratio]{#4}%
}%

\end{tabularx}
\end{tcolorbox}
}

\newcommand{\textpostentry}[3]{%
\begin{tcolorbox}[
  enhanced,
  colback=white,
  colframe=twitterblue!22,
  boxrule=0.25pt,
  arc=2pt,
  left=6pt,
  right=6pt,
  top=6pt,
  bottom=6pt,
  before skip=6pt,
  after skip=6pt
]
{\bfseries #1}\quad{\scriptsize\color{black!60}#2}\par
\vspace{3pt}
#3
\end{tcolorbox}
}

\title{SocialPersona: Benchmarking Personalized Profiling and Response with Multimodal Social-Media Context}

\newcommand{\bench}{\textsc{SocialPersona }}

\author{Qinkai Zhang$^1$, Yanyan Zhao$^1$\thanks{Corresponding author.}, Xin Lu$^1$, Yulin Hu$^1$, Pengtao Han$^1$, Bing Qin$^1$\\
  $^1$Harbin Institute of Technology \\
  \texttt{\{qkzhang, yyzhao\}@ir.hit.edu.cn}}

\begin{document}

\maketitle

\begin{abstract}
Personalized language-model assistants are often evaluated through a memory lens: can a model recall preferences users have explicitly stated in dialogue? More comprehensive personalization demands a harder capability---inferring what users care about from the multimodal traces they naturally leave behind. We introduce \bench{}, a benchmark for evaluating whether multimodal large language models (MLLMs) can recover revealed preferences from longitudinal social-media timelines and use them in dialogue. Built from longitudinal timelines of 171 everyday, non-promotional social-media users, \bench{} contains text, images, timestamps, and 2,597 human-verified preference tags across seven interest domains, separating stable interests from recent interests. It supports two tasks: constructing structured user profiles from multimodal context and generating responses aligned with inferred profiles. Experiments with proprietary and open-weight MLLMs show that models can identify broad interest domains, yet their performance drops on fine-grained and recent interests and degrades further when inferred profiles must be used to personalize dialogue. Together with evidence that text and images provide complementary preference signals, these results indicate that robust cross-modal, long-horizon user modeling remains a key challenge, and that \bench{} can help measure and advance progress toward assistants that infer and act on revealed preferences.
\end{abstract}

\section{Introduction}

\begin{figure}[t]
    \centering
    \includegraphics[width=0.5\textwidth]{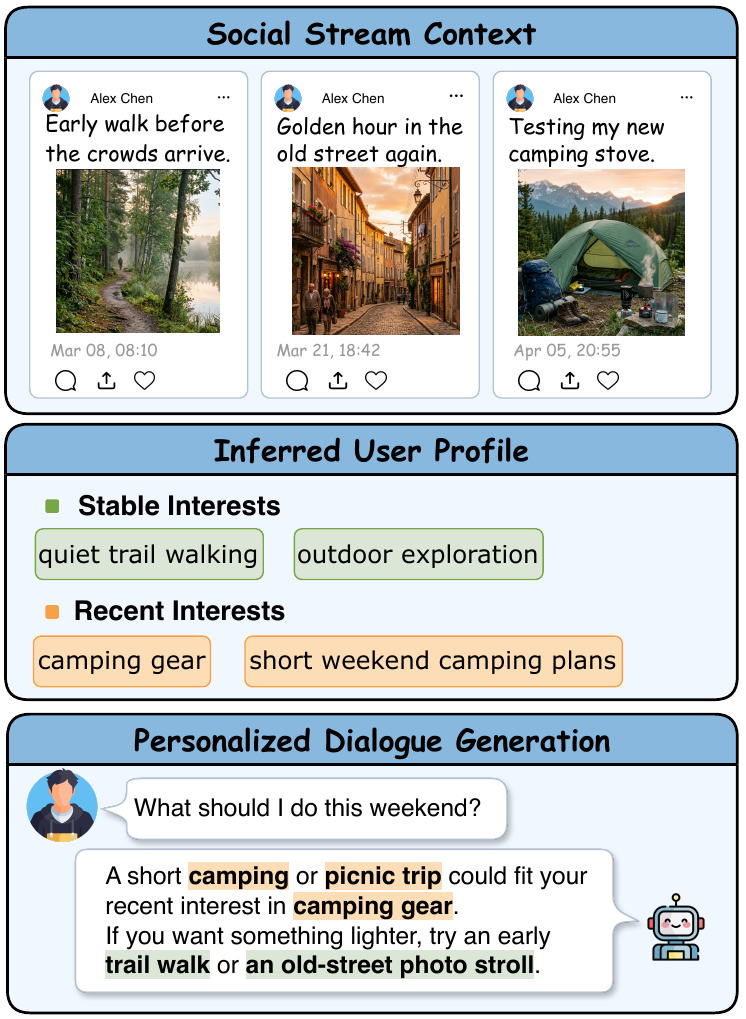}
    \caption{A user's social-media timeline provides textual, visual, and temporal evidence for stable and recent interests, which can guide personalized responses to new queries.}%
  \label{fig:motivation}
\end{figure}

Personalized assistants are increasingly expected to account for a user's long-term interests, recent activities, and implicit preferences~\citep{chen2024llmpersonalization,liu2025personalizedllms,purificato2024survey}. Existing benchmarks, however, mainly test whether models remember preferences explicitly stated in dialogue, emphasizing \emph{memory} rather than \emph{insight}. In practice, preferences are often revealed indirectly through what users create, share, photograph, discuss, and repeatedly engage with~\citep{he2023userbehavior,huang2026mempal}. Social-media timelines offer a rich source of such signals, but recovering them requires aggregating multimodal evidence over time, distinguishing stable hobbies from recent fixations, and applying the inferred profile in personalized interaction. Figure~\ref{fig:motivation} illustrates how timeline evidence can be transformed into stable and recent interests for dialogue.

Prior benchmarks mostly represent user context as dialogue-derived stated preferences~\citep{salemi2023lamp,zhao2025prefeval,jiang2025personamem,zhao2025personalens}. Although recent work incorporates longer behavioral histories~\citep{huang2026mempal}, it still relies on synthetic or structured textual logs. These settings bypass a key challenge for MLLMs: inferring user interests from noisy, unstructured, longitudinal social-media traces, where evidence is weak, distributed across posts, and often available only through images, timestamps, or cross-post patterns.

We introduce \bench{}, a benchmark for evaluating \emph{MLLM personalization from multimodal social-media context}. Built from real timelines of everyday, non-promotional users, \bench{} contains chronologically organized text, images, and timestamps. From these timelines, we construct human-validated interest profiles across seven domains: sports and outdoor activities, entertainment, gaming, food and drink, travel and city exploration, photography and creation, and pets. Each profile separates stable interests from recent interests and grounds them in supporting evidence.

\bench{} supports two evaluation settings. In \emph{profile construction}, models infer active domains and fine-grained interest tags from raw multimodal timelines. In \emph{personalized dialogue generation}, models receive social-media context with a current request and are evaluated on whether their responses align with the user's stable or recent interests. Together, these tasks test whether MLLMs can both recover implicit preferences and use them in downstream interaction.

Concretely, \bench{} contains timelines from 171 real users, with an average of 176.81 posts and 130.38 images per user. A semi-automated pipeline followed by human verification yields 2,597 preference tags grounded in textual, visual, and temporal evidence. Experiments with proprietary and open-weight MLLMs show that current models still struggle to infer fine-grained and recent interests, and to consistently use inferred profiles in personalized responses.

Our contributions are three-fold:
\begin{enumerate}
    \item We introduce a new task formulation that challenges MLLMs to infer user preferences from longitudinal, multimodal social-media behavior—aggregating sparse textual, visual, and temporal signals across long horizons—and to apply the inferred preferences in personalized dialogue generation.
    \item We construct \bench{}, a real-user benchmark with long-horizon timelines, multimodal evidence, timestamps, and human-validated profiles across seven domains, and publicly release benchmark code with a de-identified evaluation subset\footnote{\label{fn:data_release}Available at \url{https://anonymous.4open.science/r/socialpersona-6E9B}. Original images are excluded to reduce re-identification risk. Qualified researchers may request controlled access to the full benchmark for replication and follow-up studies; please contact the authors at \texttt{qkzhang@ir.hit.edu.cn} for details.}.
    \item We evaluate proprietary and open-weight MLLMs on profile construction and personalized dialogue generation, revealing gaps in cross-modal evidence aggregation and user-aligned response generation.
\end{enumerate}

\section{Related Work}
\begin{table*}[t]
    \centering
    \footnotesize
    \setlength{\tabcolsep}{4pt} 
    \renewcommand{\arraystretch}{1.2} 
    \begin{tabular}{@{} l p{4cm} c c c c c @{}}
        \toprule
        Benchmark & Context source & \makecell{Multi-\\modal} & \makecell{Real\\Data} & \makecell{Revealed\\Pref.} & \makecell{Profile\\Eval} & \makecell{Dialogue\\Eval} \\
        \midrule
        LaMP \citep{salemi2023lamp}
            & user text history & \no & \yes & \no & \no & \no \\
        PrefEval \citep{zhao2025prefeval}
            & dialogue history & \no & \no & \no & \no & \yes \\
        PERSONAMEM \citep{jiang2025personamem, jiang2025personamemv2}
            & dialogue history & \yes & \no & \no & \yes & \yes \\
        Mem-PAL \citep{huang2026mempal}
            & behavioral logs + dialogue & \no & \no & \yes & \yes & \yes \\
        ALPBench \citep{ren2026alpbench}
            & e-commerce behavior & \no & \yes & \yes & \yes & \no \\
        GISTBench \citep{fostiropoulos2026gistbench}
            & short-video engagement & \no & \no & \yes & \yes & \no \\
        \textbf{\bench{} (ours)}
            & \textbf{user social timeline} & \yes & \yes & \yes & \yes & \yes \\
        \bottomrule
    \end{tabular}
    \caption{Comparison of personalization benchmarks across context source, modality, data provenance, preference source, and evaluation target. ``Real Data'' denotes organically accumulated user-generated evidence; ``Revealed Pref.'' denotes preference signals inferred from behavioral traces. \bench{} is the only benchmark covering multimodal real-user social timelines, revealed preferences, and both profile and dialogue evaluation.}

    \label{tab:benchmark-comparison}
\end{table*}

\subsection{Personalization Benchmarks}

Recent personalization benchmarks mainly construct user context from dialogue histories or structured behavior logs. LaMP~\citep{salemi2023lamp} evaluates personalized language tasks from user-specific textual histories, while PrefEval~\citep{zhao2025prefeval}, PersonaMem~\citep{jiang2025personamem}, and PersonaLens~\citep{zhao2025personalens} study preference recognition, user memory, and personalized response generation from conversational context. More recent benchmarks move toward longer-term behavioral modeling, including Mem-PAL~\citep{huang2026mempal} for behavioral-log-grounded dialogue and ALPBench~\citep{ren2026alpbench} / GISTBench~\citep{fostiropoulos2026gistbench} for e-commerce or short-video interest inference. Agent-oriented benchmarks further extend personalization to search, web, and mobile environments~\citep{kim2025bespoke,cai2024personalwab,kim2026persona2web,yang2025fingertip,chen2026knowu}.

As summarized in Table~\ref{tab:benchmark-comparison}, \bench{} differs from prior benchmarks by combining multimodal input, real-user data, revealed-preference signals, profile evaluation, and dialogue evaluation in one setting.

\subsection{Multimodal Social-media Understanding}

Prior multimodal social-media datasets study content-level tasks such as sentiment and affect analysis~\citep{niu2016sentiment,yu2019adapting,sharma2020semeval}, sarcasm and humor detection~\citep{cai2019multi}, crisis response~\citep{alam2018crisismmd}, misinformation verification~\citep{shu2020fakenewsnet,nakamura2020fakeddit,nielsen2022mumin,mishra2022factify,yao2023mocheg}, harmful-content recognition~\citep{kiela2020hateful,lin2024goatbench}, and broad MLLM evaluation on social-networking scenarios~\citep{zhang2024somelvlm,jin2024mmsoc,guo2025snsbenchvl}. However, these benchmarks primarily label individual posts or interactions for predefined tasks. User-related signals, when included, are usually treated as demographic attributes, engagement prediction, or recommendation targets. \bench{} instead treats a user's timeline as external personalization context: models must aggregate sparse textual, visual, and temporal evidence across many posts, distinguish stable from recent interests, and generate responses aligned with the inferred profile.

\section{\bench Construction}

\subsection{Problem Setting}

We study whether MLLMs can infer and use preferences from social-media timelines. For each user $u$, a temporally ordered timeline $\mathcal{S}_u = \langle p_1, \dots, p_n \rangle$ consists of posts $p_i = (x_i, v_i, \tau_i)$ with text $x_i$, visuals $v_i$, and timestamp $\tau_i$, spanning at most 200 posts over two years.

We define profiles over seven interest domains adapted from prior preference taxonomies~\citep{zhao2025prefeval} and platform-level interest categories:
$\{$sports\_outdoor, entertainment, gaming, food\_drink, travel\_city\_exploration, photography\_creation, pets$\}$.
For each active domain, the gold profile contains stable interests (recurring patterns across the timeline), recent interests (emerging or time-local signals near the end of the observation window), and supporting evidence links retained for auditability. We exclude demographic, identity-related, health, political, and other sensitive attributes.

\begin{figure*}[t]
    \centering
    \includegraphics[width=\textwidth]{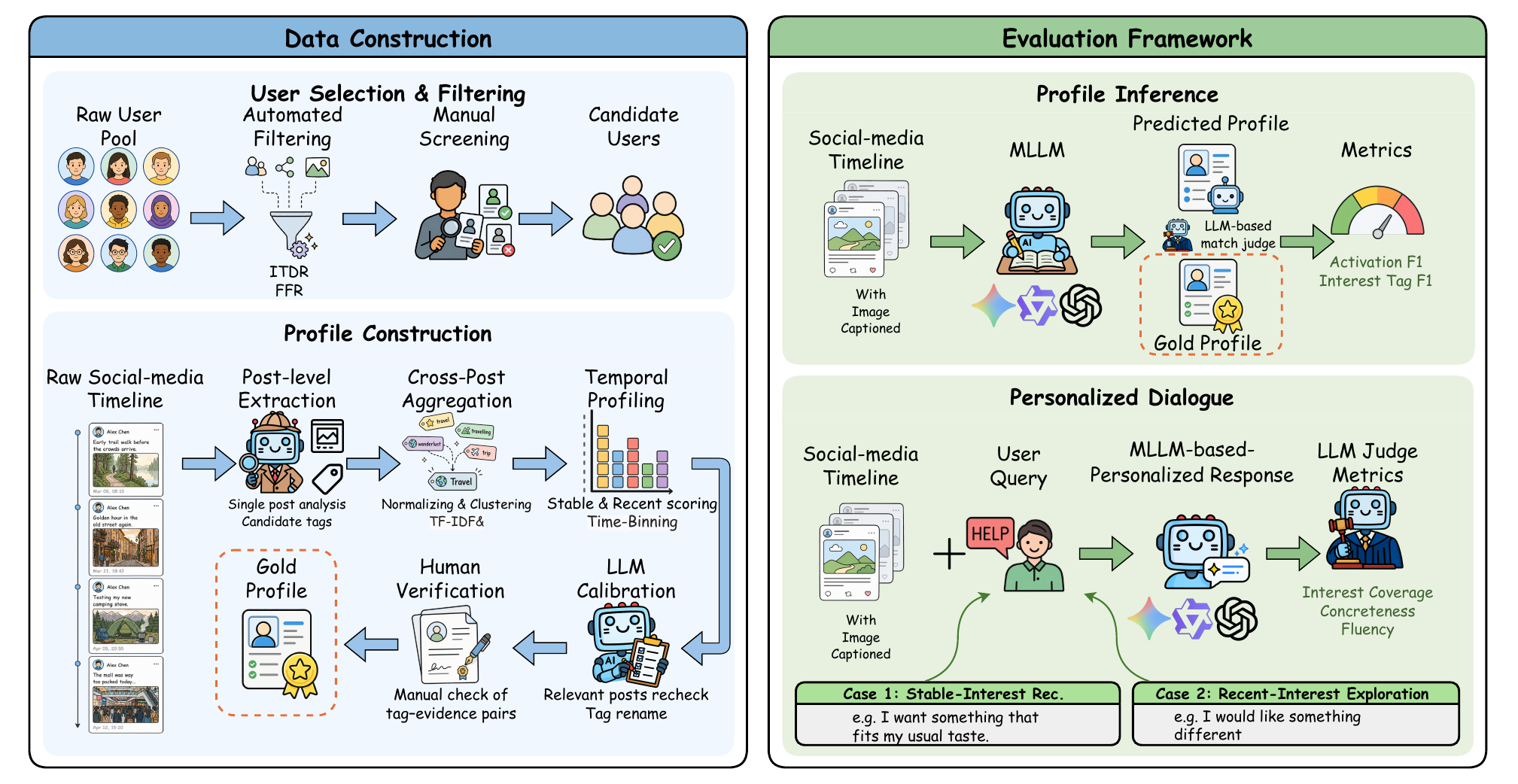}
    \caption{\textbf{Overview of SOCIALPERSONA.} \bench is constructed from real multimodal social-media timelines through user filtering, post-level interest extraction, cross-post aggregation, temporal profiling, LLM calibration, and human verification, yielding gold profiles with stable and recent interests. The benchmark evaluates MLLMs on two tasks: inferring user profiles from social media timelines, measured by domain activation and interest-tag F1, and generating personalized dialogue responses for stable-interest recommendation and recent-interest exploration, judged by interest coverage, concreteness, and fluency.}%
    \label{fig:main_image}
\end{figure*}

\bench supports two tasks. In \textbf{profile construction}, a model predicts stable and recent interest tags from the user timeline. In \textbf{personalized dialogue generation}, a model receives the timeline together with a natural user request and generates a response aligned with the user's stable or recent interests. The overall construction and evaluation pipeline is shown in Figure~\ref{fig:main_image}: \bench first converts raw social-media timelines into human-verified stable and recent interest profiles, and then evaluates whether MLLMs can recover these profiles and use them in personalized dialogue.

\subsection{User and Timeline Collection}
\label{subsec:user_selection}

We construct \bench{} from real social-media timelines of \emph{long-tail organic users}, rather than celebrities, brand accounts, or highly curated public profiles. This design choice is intended to capture relatively natural, self-expressive preference traces instead of broadcast-oriented content. Starting from 8{,}000 candidate accounts, we apply automatic filters based on follower count, follower--followee ratio, and image trace density, retaining accounts with 5--5{,}000 followers, FFR in $[0.5,2]$, and ITDR $\geq 0.3$. These filters remove extremely sparse, highly public, or insufficiently multimodal accounts while reducing the presence of broadcaster-style users~\citep{oshimo2022follower,leavitt2009influentials}. We then manually inspect the remaining accounts to exclude commercial, repost-heavy, or otherwise low-quality cases, resulting in 250 candidate users. Detailed definitions of FFR, ITDR, and the manual filtering criteria are provided in Appendix~\ref{app:user_filtering}.

For each selected user, we collect up to 200 posts from the most recent two-year window. Each post is stored as a structured multimodal record containing its timestamp, textual content, hashtags, URLs, and attached visual content, including images or video cover frames. As the original posts come from users across multiple countries and languages, we standardize all textual content by translating it into English, thereby enabling consistent profile construction and evaluation. After profile construction and verification, we further remove users with fewer than three active interest domains, as such profiles provide insufficient personalization signals for reliable evaluation. This yields the final benchmark of 171 users.

\subsection{Gold Profile Construction}
\label{subsec:profile_construction}

Given each user's multimodal timeline, we construct gold profiles with an LLM-assisted but human-verified pipeline. First, we use \texttt{Gemini-3-Flash}~\citep{googledeepmind2025gemini3flash,google2026gemini3flashapi} to perform conservative post-level extraction from the original text and visual content. The extractor is instructed to identify only observable, evidence-grounded interest signals, record modality attribution, and avoid demographic, identity-related, or speculative claims.

Second, we aggregate post-level candidates across each timeline. Near-duplicate posts are down-weighted, semantically equivalent tags are merged into canonical interests, and each user-domain pair is represented as an evidence pack. Third, we compute preliminary stable and recent assignments using duplicate-adjusted support, temporal dispersion, recency, and confidence. Stable interests require repeated support across the timeline, while recent interests emphasize evidence concentrated in the most recent 90 days.

Fourth, we apply an LLM-based calibration stage using \texttt{Gemini-3.1-Pro}~\citep{googledeepmind2026gemini31proeval,google2026gemini31proapi}. This stage checks the aggregated evidence packs, canonical candidates, scores, and preliminary temporal buckets for weak support, over-generalization, speculative labels, and bucket errors. Finally, trained annotators manually verify each calibrated profile against its supporting posts. Accepted interests must be concrete, domain-appropriate, sufficiently supported, and assigned to the correct temporal bucket.

\paragraph{Quality assurance.}
Five trained annotators independently verified each user profile, with disagreements resolved by an additional adjudicator. The process required approximately 350 annotator-hours. On a 40-user overlap subset, inter-annotator agreement reached Krippendorff's $\alpha=0.72$ for tag acceptance and $\alpha=0.63$ for stable/recent bucket assignment. Human verification modified about 12\% of pipeline-proposed tags: 8\% removed for insufficient evidence, 3\% refined to more concrete labels, and 1\% added as missed but supported interests, confirming that human oversight is essential for benchmark quality.

\section{Evaluation and Experiments}

\subsection{Evaluation Setup}
\label{subsec:evaluation}

All experiments use a fixed 100-user subset to ensure comparable cost and conditions. We evaluate profile construction and personalized dialogue generation using a timeline representation consisting of post text, image captions, and timestamps.

\paragraph{Evaluated models.}
Our main model suite is designed to cover both proprietary and open-weight MLLMs. The initial suite contains six models: Gemini-2.5-Flash~\citep{comanici2025gemini25}, GPT-4o-mini~\citep{openai2024gpt4omini}, GPT-5.4~\citep{openai2026gpt54thinking}, Qwen2.5-VL-7B-Instruct~\citep{bai2025qwen25vl}, Qwen3-VL-8B-Instruct~\citep{bai2025qwen3vl}, and Qwen3.5-35B-A3B~\citep{qwen2026qwen35a3b}.

\paragraph{Profile construction evaluation.}
Given $\widetilde{\mathcal{S}}_u^{M}$, each method predicts stable and recent interests over seven domains. A domain is active if it has at least one gold interest, and predicted active if the method outputs any interest in it. For interest-tag recovery, we evaluate each user, domain, and bucket $b \in \{\mathrm{stable}, \mathrm{recent}\}$ using normalized exact match, followed by \texttt{o3}~\citep{openai2025o3o4mini}-based semantic matching for unmatched tags as in Appendix~\ref{app:profile_match_prompt}. Matches are one-to-one and define true positives; unmatched predictions and gold tags are false positives and false negatives.

\paragraph{Personalized dialogue evaluation.}
The dialogue task evaluates whether a model can generate personalized but not over-personalized recommendations from social-media-derived user information. We evaluate four dialogue input settings. The first is a \emph{timeline-conditioned} setting, where the model receives the user's post text, image captions, and timestamps directly. The remaining three are two-stage \emph{profile-conditioned} settings: the model first constructs a profile using one of the three profile-construction settings described below, and then generates a dialogue response using only that generated profile as personalization context. These settings test whether a model-generated profile can serve as an effective intermediate memory representation, rather than only being evaluated as a structured prediction.

For each dialogue setting, we consider two intents: stable-interest recommendation, which targets the user’s long-term preferences, and recent-interest exploration, which introduces mildly novel yet personally suitable items. For each setting, we prepare ten natural requests and randomly sample one for each user--setting instance; the full pools are given in Appendix~\ref{app:dialogue_query_pool}. We use \texttt{GPT-5.5}~\citep{openai2026gpt55} and \texttt{Qwen3.7-Max}~\citep{qwen2026qwen37} as judges, following the prompt in Appendix~\ref{app:dialogue_judge_prompt}. Given the gold profile, the judge scores each response on \emph{interest coverage}, \emph{concreteness}, and \emph{fluency}.

\subsection{Main Results: Profile Construction Across Settings and Models}
\label{subsec:profiling_results}

We evaluate profile construction across three settings and six models (Table~\ref{tab:profiling_results}), separating model capability from the strategy used to handle long timelines.

\paragraph{Profile-construction settings.}
\emph{Direct} feeds the full timeline (text, image captions, timestamps, up to 200 posts) in one pass. \emph{Hierarchical} splits the timeline into 20-post chunks, summarizes each independently, then aggregates chunk summaries into the final profile. \emph{Extractive--abstractive} proceeds in two stages as detailed in Appendix~\ref{app:extractive_prompts}. First (\emph{extraction}), the LLM is prompted to select up to $K=12$ representative posts per domain, guided by relevance, specificity, recurrence, recency, and multimodal grounding. Second (\emph{abstractive synthesis}), the LLM generates the profile using only the selected posts as evidence, with no access to the original full timeline.

\renewcommand{\tabularxcolumn}[1]{m{#1}} 
\definecolor{tabgray}{gray}{0.95} 

\begin{table*}[t]
\centering
\footnotesize 
\setlength{\tabcolsep}{6pt} 
\renewcommand{\arraystretch}{1.15}
\begin{tabular}{l *{6}{c}}
\toprule
\textbf{Model} & \textbf{Active F1} & \textbf{Int. Prec.} & \textbf{Int. Rec.} & \textbf{Int. F1} & \textbf{Stable F1} & \textbf{Recent F1} \\
\midrule

\rowcolor{tabgray} \multicolumn{7}{c}{\textbf{Direct}} \\ 
\midrule
Gemini-2.5-Flash       & \best{0.8663} & 0.5323 & 0.2379 & 0.3288 & \best{0.3516} & 0.0219 \\
GPT-4o-mini            & 0.7454 & 0.4627 & 0.1992 & 0.2785 & 0.1832 & \best{0.0858} \\
GPT-5.4                & 0.8383 & \best{0.6038} & \best{0.2307} & \best{0.3338} & 0.3405 & 0.0687 \\
Qwen2.5-VL-7B-Instruct & 0.0647 & 0.4643 & 0.0085 & 0.0167 & 0.0228 & 0.0000 \\
Qwen3-VL-8B-Instruct   & 0.5969 & 0.3636 & 0.1625 & 0.2246 & 0.2148 & 0.0390 \\
Qwen3.5-35B-A3B        & 0.7875 & 0.5490 & 0.1763 & 0.2669 & 0.2989 & 0.0133 \\
\midrule

\rowcolor{tabgray} \multicolumn{7}{c}{\textbf{Hierarchical}} \\ 
\midrule
Gemini-2.5-Flash       & 0.7337 & \best{0.5556} & 0.0426 & 0.0791 & 0.0663 & 0.0300 \\
GPT-4o-mini            & 0.8260 & 0.4201 & 0.2687 & 0.3277 & 0.3283 & 0.0653 \\
GPT-5.4                & \best{0.8295} & 0.5121 & \best{0.3480} & \best{0.4144} & \best{0.3932} & \best{0.1202} \\
Qwen2.5-VL-7B-Instruct & 0.7395 & 0.3818 & 0.1153 & 0.1772 & 0.1549 & 0.0291 \\
Qwen3-VL-8B-Instruct   & 0.8038 & 0.3971 & 0.2706 & 0.3219 & 0.2802 & 0.1129 \\
Qwen3.5-35B-A3B        & 0.8066 & 0.4556 & 0.2320 & 0.3074 & 0.3409 & 0.0679 \\
\midrule
 
\rowcolor{tabgray} \multicolumn{7}{c}{\textbf{Extractive--abstractive}} \\ 
\midrule
Gemini-2.5-Flash       & 0.2971 & \best{0.5930} & 0.0334 & 0.0633 & 0.0514 & 0.0270 \\
GPT-4o-mini            & 0.8249 & 0.4297 & 0.1442 & 0.2159 & 0.0195 & 0.0746 \\
GPT-5.4                & 0.8428 & 0.4500 & \best{0.3211} & \best{0.3748} & \best{0.3388} & \best{0.1102} \\
Qwen2.5-VL-7B-Instruct & 0.3735 & 0.2717 & 0.0164 & 0.0309 & 0.0053 & 0.0123 \\
Qwen3-VL-8B-Instruct   & 0.5247 & 0.3259 & 0.0478 & 0.0834 & 0.0640 & 0.0188 \\
Qwen3.5-35B-A3B        & \best{0.8599} & 0.3829 & 0.2058 & 0.2677 & 0.2349 & 0.0906 \\
\bottomrule
\end{tabular}
\caption{Profile construction results across three construction settings and six models. Best results within each setting are \best{bolded}.}
\label{tab:profiling_results}
\end{table*}

\renewcommand{\tabularxcolumn}[1]{m{#1}} 
\definecolor{tabgray}{gray}{0.95} 

\begin{table*}[t]
\centering
\footnotesize 
\renewcommand{\arraystretch}{1.15} 

\begin{tabular}{l cc @{\hspace{2em}} cc @{\hspace{2em}} cc @{\hspace{2em}} cc}
\toprule
\multirow{2}{*}[-0.8ex]{\textbf{Model}} & \multicolumn{2}{c}{\textbf{Cov.}} & \multicolumn{2}{c}{\textbf{Conc.}} & \multicolumn{2}{c}{\textbf{Flu.}} & \multicolumn{2}{c}{\textbf{Avg.}} \\
\cmidrule(lr){2-3} \cmidrule(lr){4-5} \cmidrule(lr){6-7} \cmidrule(lr){8-9}
& \textbf{GPT-5.5} & \textbf{Qwen3.7} & \textbf{GPT-5.5} & \textbf{Qwen3.7} & \textbf{GPT-5.5} & \textbf{Qwen3.7} & \textbf{GPT-5.5} & \textbf{Qwen3.7} \\
\midrule

\rowcolor{tabgray} \multicolumn{9}{c}{\textbf{Timeline}} \\
\midrule
Gemini-2.5-Flash       & 1.8800 & 2.4200 & 3.1700 & 3.1200 & 3.8400 & 4.2000 & 2.9633 & 3.2467 \\
GPT-4o-mini            & 1.9700 & 2.2500 & 3.5100 & 2.8700 & 3.9200 & 3.8900 & 3.1333 & 3.0033 \\
GPT-5.4                & \textbf{2.3600} & \textbf{2.8100} & \textbf{4.0200} & \textbf{4.1600} & \textbf{4.3200} & \textbf{4.7300} & \textbf{3.5667} & \textbf{3.9000} \\
Qwen2.5-VL-7B-Instruct & 1.6300 & 1.7800 & 3.0400 & 2.4000 & 2.9700 & 2.7000 & 2.5467 & 2.2933 \\
Qwen3-VL-8B-Instruct   & 2.2200 & 2.5500 & 3.7300 & 3.5200 & 3.9600 & 4.2300 & 3.3033 & 3.4333 \\
Qwen3.5-35B-A3B        & 1.6300 & 2.0300 & 3.4200 & 3.1200 & 3.8400 & 3.9900 & 2.9633 & 3.0467 \\
\midrule

\rowcolor{tabgray} \multicolumn{9}{c}{\textbf{Direct}} \\
\midrule
Gemini-2.5-Flash       & 1.6400 & 2.0600 & 2.6300 & 2.2800 & 3.7600 & 3.8700 & 2.6767 & 2.7367 \\
GPT-4o-mini            & 1.9500 & 2.2300 & 3.2600 & 2.7400 & 3.8200 & 3.9600 & 3.0100 & 2.9767 \\
GPT-5.4                & 2.5400 & \textbf{3.0400} & \textbf{3.9200} & \textbf{3.9400} & \textbf{4.1900} & \textbf{4.6500} & \textbf{3.5500} & \textbf{3.8767} \\
Qwen2.5-VL-7B-Instruct & 0.8800 & 1.0800 & 2.4700 & 1.8300 & 3.1500 & 2.9500 & 2.1667 & 1.9533 \\
Qwen3-VL-8B-Instruct   & \textbf{2.5600} & 2.8000 & 3.5700 & 3.2400 & 4.0300 & 4.2900 & 3.3867 & 3.4433 \\
Qwen3.5-35B-A3B        & 1.9300 & 2.3300 & 3.0200 & 2.6500 & 3.7900 & 4.0100 & 2.9133 & 2.9967 \\
\midrule

\rowcolor{tabgray} \multicolumn{9}{c}{\textbf{Hierarchical}} \\
\midrule
Gemini-2.5-Flash       & 1.4000 & 1.6100 & 2.6300 & 2.1500 & 3.6800 & 3.9600 & 2.5700 & 2.5733 \\
GPT-4o-mini            & 2.0300 & 2.2100 & 3.2700 & 2.6100 & 3.8900 & 3.9100 & 3.0633 & 2.9100 \\
GPT-5.4                & \textbf{2.5000} & \textbf{2.8900} & \textbf{3.8100} & \textbf{3.7200} & \textbf{4.1400} & \textbf{4.5000} & \textbf{3.4833} & \textbf{3.7033} \\
Qwen2.5-VL-7B-Instruct & 1.7100 & 1.7700 & 2.9700 & 2.2200 & 3.3700 & 3.2000 & 2.6833 & 2.3967 \\
Qwen3-VL-8B-Instruct   & 2.0700 & 2.2300 & 3.2900 & 2.9600 & 3.9400 & 3.9600 & 3.1000 & 3.0500 \\
Qwen3.5-35B-A3B        & 1.8800 & 2.2200 & 3.1200 & 2.6900 & 3.7500 & 3.9300 & 2.9167 & 2.9467 \\
\midrule

\rowcolor{tabgray} \multicolumn{9}{c}{\textbf{Extractive--abstractive}} \\  
\midrule
Gemini-2.5-Flash       & 0.9300 & 1.2200 & 2.4800 & 2.0400 & 3.5800 & 3.8500 & 2.3300 & 2.3700 \\
GPT-4o-mini            & 1.6200 & 2.0100 & 3.2000 & 2.5400 & 3.8400 & 3.9000 & 2.8867 & 2.8167 \\
GPT-5.4                & \textbf{2.5450} & \textbf{2.9700} & \textbf{4.0350} & \textbf{3.8900} & \textbf{4.2650} & \textbf{4.6200} & \textbf{3.6150} & \textbf{3.8267} \\
Qwen2.5-VL-7B-Instruct & 1.2400 & 1.3700 & 2.6700 & 1.9400 & 3.2500 & 3.0300 & 2.3867 & 2.1133 \\
Qwen3-VL-8B-Instruct   & 1.5300 & 1.6800 & 3.0800 & 2.6400 & 3.9700 & 4.0000 & 2.8600 & 2.7733 \\
Qwen3.5-35B-A3B        & 1.9300 & 2.1900 & 3.2000 & 2.7600 & 3.7900 & 3.9800 & 2.9733 & 2.9767 \\
\bottomrule
\end{tabular}
\caption{Personalized dialogue generation results. Profile-conditioned settings use generated profiles from the corresponding construction setting in Table~\ref{tab:profiling_results}.}
\label{tab:dialogue_results}
\end{table*}

\paragraph{Over-generalization trumps recall.}
The dominant failure mode across all models and settings is systematic over-generalization: models reduce 4--7 gold interest tags to 1--2 broad categories per active domain. Averaged over the direct setting, gold profiles contain 3.4 tags per active domain, while models predict only 1.5. This gap is consistent across all evaluated models and represents a fundamental limitation of current MLLMs in fine-grained preference elicitation from social-media evidence. A prompt analysis in Appendix~\ref{app:prompt_analysis} confirms that this over-generalization is not an artifact of the evaluation prompt: removing the conservative instruction to ``prefer fewer, broader tags'' does not materially change Interest F1, indicating that the recall gap reflects genuine model limitations rather than benchmark design.

\paragraph{Domain blindness follows evidence modality.}
Domain detection varies with evidence modality. Text-dispersed interests, such as music habits, city exploration, and meals, are often missed: in Appendix~\ref{app:case_studies}, most models incorrectly mark text-heavy travel and food-drink domains as inactive. In contrast, visually concentrated domains, such as pets and gaming screenshots, obtain the highest F1 scores across settings (Table~\ref{tab:per_domain}). This asymmetry indicates a reliance on visual salience for domain activation: models reliably activate domains when images directly show relevant content, but often default to inactive when evidence is diffuse and textual. Even GPT-5.4, despite the best overall tag F1, misses 12.5\% of gold-active domains under direct profiling.

\paragraph{Recent-interest blindness is a temporal reasoning gap.}
Stable F1 exceeds Recent F1 across every model, setting, and input mode. This is not simply because recent interests are ``harder''---it reflects a fundamental limitation in how models process timelines. Distinguishing an interest that recurs across 18~months (stable) from one concentrated in the most recent 90~days (recent) requires tracking temporal dispersion across posts. Current models receive posts as a flattened sequence; timestamps are present in the input but models show no evidence of computing temporal distribution patterns from them. Gemini-2.5-Flash and Qwen3.5-35B-A3B assign nearly all predicted tags to the stable bucket regardless of the actual temporal evidence, while GPT-4o-mini---the only model with non-trivial Recent F1 (0.086)---distributes tags more evenly between buckets but without temporal alignment, suggesting its higher Recent F1 reflects a flatter prior over buckets rather than genuine temporal discrimination.

\paragraph{Hierarchical profiling amplifies model-specific behaviors.}
Hierarchical profiling improves Interest F1 for strong models (GPT-5.4, GPT-4o-mini) but causes conservative models to collapse, as sparse chunk-level summaries leave the aggregation step with insufficient evidence. The same pattern recurs in the extractive setting: capable models benefit from the two-stage decomposition while weaker models degrade further. Detailed per-model breakdowns are provided in Appendix~\ref{app:case_studies}.

\subsection{Main Results: Personalized Dialogue Generation}
\label{subsec:dialogue_results}

Table~\ref{tab:dialogue_results} reports dialogue results across four input settings and six models. The \emph{timeline-conditioned} setting provides the raw timeline directly. The other three are \emph{profile-conditioned}: the model first constructs a profile using one of the three settings from Table~\ref{tab:profiling_results}, then generates a response using only that profile. We report three judge-scored dimensions (0--5 scale) and their unweighted average $\mathrm{Avg} = (\mathrm{Coverage} + \mathrm{Concreteness} + \mathrm{Fluency}) / 3$.

\paragraph{Profile conditioning is a double-edged filter.}
Profile conditioning filters timeline noise but also propagates profiling errors. When generated profiles are sparse, personalization coverage and concreteness drop; when profiles are richer, the intermediate profile can improve dialogue quality. For example, Gemini-2.5-Flash loses coverage under profile conditioning, while GPT-5.4 slightly improves coverage, suggesting that profile quality determines whether the profile acts as a useful memory representation or an information bottleneck. The profile--dialogue correlation analysis (Section~\ref{subsec:additional_analyses}) confirms this dependency quantitatively, with 16 of 18 Interest~F1--Avg.\ correlations positive and stronger under the tighter information bottleneck of hierarchical and extractive profiling. Fluency remains relatively stable across settings and models (3.0--4.3 for most), indicating that the core challenge is not generating readable text but producing responses that engage the correct interests.

\paragraph{Human--LLM agreement.}
A 30-response human agreement study confirms that the three dialogue dimensions are reliably judgeable and that both LLM judges track human rankings. Inter-annotator agreement is substantial for fluency, concreteness, and coverage ($\alpha=0.79/0.71/0.65$), and GPT-5.5 correlates most strongly with human mean scores ($\rho=0.76/0.65/0.58$, all $p<0.01$). Full sampling and annotation details are in Appendix~\ref{app:human_agreement}.

\subsection{Additional Analyses}
\label{subsec:additional_analyses}

\paragraph{Input-modality analysis.}
Input modality analysis on Qwen3.5-35B-A3B and GPT-4o-mini shows that text provides the main profiling signal, while image captions supply complementary tacit cues: adding captions raises GPT-4o-mini's Active F1 from 0.503 to 0.737. Timestamps further improve Active F1 but leave Tag F1 largely unchanged, suggesting that temporal structure helps domain detection more than fine-grained tag recovery.

\paragraph{Per-domain difficulty.} Per-domain breakdown (Table~\ref{tab:per_domain}) confirms that domains with visually distinctive evidence (pets, gaming) are consistently easier across settings, while domains requiring synthesis of dispersed evidence (travel, entertainment) are hardest.

\paragraph{Profile--dialogue correlation.} The two-stage design enables a direct test of whether profile quality translates to dialogue quality. Per-user Spearman rank correlations show that profile Interest F1 positively correlates with dialogue quality: 16 of 18 Interest F1--Avg.\ correlations are positive (6 significant at $p<0.05$), with stronger correlations under Hierarchical and Extractive profiling, where the profile acts as a tighter information bottleneck. Interest F1 correlates most strongly with coverage (15 of 18 pairs significant), confirming that better profiling primarily improves interest engagement.

\section{Conclusion}

We introduced \bench, a benchmark for personalized user profiling and response generation from multimodal social-media timelines, built from real social-media user data with human-validated interest profiles across seven domains. Our experiments reveal three central findings. First, current models systematically over-generalize, collapsing specific interests into broad categories---a genuine capability gap, not an artifact of conservative evaluation instructions (Appendix~\ref{app:prompt_analysis}). Second, models exhibit pronounced modality asymmetry: visually salient domains are reliably detected while text-distributed domains are frequently missed, and this bias propagates into dialogue. Third, distinguishing stable from recent interests remains beyond current capabilities; models lack the cross-post temporal reasoning needed to separate persistent patterns from emerging ones. The profile--dialogue correlation validates the two-stage design but reveals that sparse profiles compound these failures downstream.

\section*{Limitations}

Our evaluation uses image captions rather than raw images because full user timelines can contain hundreds of images, making raw-image evaluation difficult under heterogeneous API limits, visual-token budgets, and upload interfaces. Captions provide a unified cross-model input format but may lose pixel-level details. The modality analysis shows that captions still add substantial signal alongside text for the interest categories studied here.

\section*{Ethical Considerations}

\bench{} is designed as an aggregate benchmark for studying personalized modeling from publicly accessible social-media content under a restricted research protocol. Annotation and evaluation are limited to non-sensitive, evidence-grounded interests, and explicitly avoid demographic, identity-related, health, political, or other sensitive inferences. We release only a de-identified text-plus-caption subset, exclude original images, and provide the full benchmark only through controlled research access. The benchmark is intended solely for aggregate model evaluation and research purposes, and must not be used for identification, surveillance, targeting, or consequential decision-making.

\bibliography{refs}

\appendix

\section{Prompt Templates}
\label{app:prompts}

\subsection{Single-Post Analysis Prompt}
\label{app:single_post_prompt}

\begin{promptbox}{Single-Post Analysis --- System Prompt}{systemblue}
You are an information extraction model for user-interest profiling from a single social media post.

Your job is to extract only observable, evidence-grounded interest signals from the post.

Fixed domains:
1. sports_outdoor: sports participation, exercise, fitness routines, hiking, running, cycling, camping, outdoor recreation, and active-use sports gear.
2. entertainment: movies, TV, music, concerts, books/comics/anime, celebrities, and media consumption. Exclude gaming unless explicitly about games.
3. gaming: video games, gaming hardware/platforms, esports, game fandom, game streaming, and playing/watching games.
4. food_drink: cooking, meals, restaurants, cafes, recipes, coffee, tea, cocktails, and other food/drink consumption or creation.
5. travel_city_exploration: trips, flights, hotels, cities, neighborhoods, sightseeing, landmarks, museums, and city walks/exploration.
6. photography_creation: taking photos, cameras, lenses, editing, visual creation, making images/videos/artworks. Do not assign this domain for a scenic image alone unless the post clearly signals photographing, editing, or creating.
7. pets: pets, pet ownership, pet care, dogs, cats, training, grooming, adoption, veterinary care, pet products, and spending time with companion animals. Exclude wildlife or general nature content unless the post is clearly about personal pets or pet care.

General rules:
- Use only explicit evidence from the provided post package and any attached image inputs: text, hashtags, mentions, URL domains, visible metadata, gate result, and optional media summary.
- In each evidence item, `source` must be either `text` or `visual`.
- Do not infer sensitive or demographic attributes.
- A post may map to zero, one, or multiple domains, but do not over-assign.
- is_noise can only be true when the gate result explicitly indicates too_little_content or meme_image_only.
- Tags must be short, reusable, canonical phrases in snake_case.
- Prefer specific tags such as trail_running, cold_brew_coffee, city_walks, concert_attendance, dog_care.
- Avoid generic tags such as sports, entertainment, lifestyle, fun, daily_life.
- Each tag must be directly supported by evidence in the input.
- Confidence values must be between 0 and 1.
- Typical output should contain at most 3 domains and 1-4 tags per domain.

Allowed tag_type values:
- activity
- preference
- subject
- place
- object
- routine
- relation

Return strict JSON only with this schema:
{
  "schema_version": "single_post_profile_v1",
  "is_noise": boolean,
  "should_use_for_profile": boolean,
  "noise_reason": "too_little_content" | "meme_image_only" | null,
  "post_signal_strength": number,
  "domains": [
    {
      "domain": string,
      "confidence": number,
      "evidence": [
        {"source": "text" | "visual", "value": string}
      ],
      "tags": [
        {"tag": string, "confidence": number, "tag_type": string}
      ]
    }
  ],
  "uncertainty_note": string or null
}
\end{promptbox}

\begin{promptbox}{Single-Post Analysis --- User Prompt}{usergreen}
Analyze the following post for user-interest profiling.

{post_package_json}

Instructions:
- Respect the gate result. If gate_result.is_noise is true, keep the output as noise and do not invent domains.
- If there is image media and the gate did not mark it as meme_image_only, do not mark the post as noise solely because the text is short.
- Only assign a domain when evidence is explicit.
- Use `source=text` for textual cues and `source=visual` for image-only cues.
- Keep the output grounded and conservative.
- Return JSON only.
\end{promptbox}

\subsection{Profile Evaluation Model Prompt}
\label{app:profile_eval_model_prompt}

\begin{promptbox}{Profile Evaluation --- System Prompt}{systemblue}
You are evaluating a user's domain-level interests from social-media posts.

Your job is to decide whether a domain is active, and if so, extract only the clearly supported interest tags.

Task:
- Decide whether the domain status is active or inactive.
- If active, extract the supported interest tags.
- If inactive, abstain cleanly: no interest tags.

Definitions:
- stable_interest_tags: recurring, reinforced, or stable interests supported across multiple posts or over time.
- recent_interest_tags: newer, narrower, or more time-local interests that are supported but not yet stable.

Rules for active domains:
- Return only clearly supported tags. Do not try to fill a quota.
- Most active domains should have only 1-2 reliable tags in total.
- Return more than 2 tags only when the evidence is unusually strong and the tags are clearly distinct.
- It is valid to return zero stable_interest_tags.
- It is valid to return zero recent_interest_tags.
- Each tag must be a short natural-language label that reflects a user interest theme, not a raw keyword list, hashtag list, named entity list, or one-off event.
- Prefer fewer, broader tags that capture the user's main tendencies in this domain.
- Do not produce near-duplicate tags across stable and recent buckets.
- If two candidate tags largely overlap, keep only the broader or better-supported one.
- Do not split one core interest into both stable and recent tags unless the recent tag adds a clearly distinct recent focus.
- Do not introduce unsupported themes, motivations, personality traits, or lifestyle claims.

Rules for inactive domains:
- stable_interest_tags must be [].
- recent_interest_tags must be [].

General rules:
- If the evidence is weak, sparse, one-off, or not clearly attributable to user preference, prefer inactive.
- If uncertain, choose fewer tags and keep the output conservative.
- Return strict JSON only.

Schema:
{
  "domain": string,
  "status": "active" | "inactive",
  "stable_interest_tags": [string],
  "recent_interest_tags": [string]
}
\end{promptbox}

\begin{promptbox}{Profile Evaluation --- User Prompt}{usergreen}
Evaluate user profile signal for one domain.

DOMAIN: {domain}
DOMAIN_DEFINITION: {domain_definition}
OUTPUT_STATUS_OPTIONS: active or inactive

POSTS_CONTEXT:
{posts_context}

Instructions:
- Read the posts and decide whether this domain contains a reliable user-interest signal.
- If active, extract only the clearly supported interest tags for this domain.
- Use stable_interest_tags for recurring or stable interests.
- Use recent_interest_tags for newer or more time-local interests.
- Return only clearly supported tags. Do not try to fill a quota.
- Most active domains should have only 1-2 reliable tags in total.
- Return more than 2 tags only when the evidence is unusually strong and the tags are clearly distinct.
- It is valid to return zero stable_interest_tags.
- It is valid to return zero recent_interest_tags.
- Use short natural-language labels that describe user interest themes rather than raw hashtags, named entities, or one-off events.
- Prefer fewer, broader tags that summarize the user's main tendencies in this domain.
- Do not produce near-duplicate tags across stable and recent buckets.
- If two candidate tags largely overlap, keep only the broader or better-supported one.
- Do not split one core interest into both stable and recent tags unless the recent tag adds a clearly distinct recent focus.
- If the evidence is weak, sparse, one-off, or not clearly attributable to user preference, prefer inactive.
- If inactive, return empty tag lists.
- Return JSON only.
\end{promptbox}

\subsection{Profile Evaluation Matching Judge Prompt}
\label{app:profile_match_prompt}

\begin{promptbox}{Profile Matching Judge --- System Prompt}{judgepurple}
You are an evaluator for domain-level user-interest anchor labels.

Given one gold anchor label and one predicted anchor label within the same domain,
decide whether they describe the same core user interest.

Judgment rules:
1. Return true only when both labels express the same core interest theme.
2. A wording rewrite, synonym, parent-child phrasing, or broad-vs-specific phrasing can be true when both labels clearly point to the same underlying user-interest cluster in this domain.
3. Examples that should usually be true: "hiking" vs "outdoor recreation", "coffee" vs "coffee culture", "anime" vs "anime fandom".
4. Examples that should usually be false: sibling interests that share only the same domain, such as "basketball" vs "camping", or labels with clearly different focus.
5. Be conservative. Return strict JSON only.
\end{promptbox}

\begin{promptbox}{Profile Matching Judge --- User Prompt}{judgepurple}
Task ID: {task_id}

Domain: {domain}
Domain definition: {domain_definition}

Gold anchor label:
{gold_label}

Predicted anchor label:
{pred_label}

Question:
Do these two labels describe the same core user interest in this domain?
Return strict JSON only.
\end{promptbox}

\subsection{Dialogue Generation Prompt}
\label{app:dialogue_gen_prompt}

\begin{promptbox}{Dialogue Generation --- System Prompts}{systemblue}
[System Prompt -- Full-context mode]

You are a concise personalization assistant.

Use only the user's posts and image captions that are provided in the context.
Give one natural, practical recommendation that fits the current user request.
Stay close to supported interests and do not invent demographics or hidden motives.
When images reveal a hobby, object, routine, food, place, or activity that the text alone would not make obvious, it is valid and desirable to use that signal.
Do not mention posts, evidence, pipelines, scoring, benchmarks, profiles, or observation windows unless the user explicitly asks.

[System Prompt -- Profile-only mode]

You are a concise personalization assistant.

Use only the user's profile information that is provided in the context.
Give one natural, practical recommendation that fits the current user request.
Stay close to supported interests and do not invent demographics or hidden motives.
Do not mention evidence, pipelines, scoring, benchmarks, profiles, or observation windows unless the user explicitly asks.
\end{promptbox}

\begin{promptbox}{Dialogue Generation --- User Prompt Templates}{usergreen}
[User Prompt Template -- Full-context mode]

PERSONALIZATION RULES:
- Use only the provided posts, image captions, and images.
- Answer the user's request naturally.
- Give one concrete recommendation, not a long list.
- Do not mention evidence, post indices, profiles, pipelines, scoring, or observation windows.

USER POSTS AND IMAGE CONTEXT:
{context}

USER REQUEST:
{user_prompt}

[User Prompt Template -- Profile-only mode]

PERSONALIZATION RULES:
- Use only the provided user profile information.
- Answer the user's request naturally.
- Give one concrete recommendation, not a long list.
- Do not mention evidence, post indices, profiles, pipelines, scoring, or observation windows.

USER PROFILE:
{context}

USER REQUEST:
{user_prompt}
\end{promptbox}

\subsection{Dialogue User Request Pool}
\label{app:dialogue_query_pool}
For dialogue evaluation, we sample one user request from the corresponding setting-specific pool. The stable-interest pool targets stable preference use, while the recent-interest pool targets recent exploration that remains compatible with the user's stable preferences.

\paragraph{Stable-interest recommendation requests.}
\begin{enumerate}
    \item I want something that fits my usual taste. Could you recommend one option for me?
    \item Could you suggest one thing I'd probably enjoy based on what I usually like?
    \item I'm looking for a recommendation that feels very me. What's one good option?
    \item Choose one option that fits what I've liked for a while.
    \item I want a safe choice that matches my usual preferences. What should I try?
    \item Recommend one activity or item that seems close to my regular taste.
    \item Based on what I tend to enjoy, what is one practical suggestion?
    \item I'm not trying to branch out today; give me one recommendation that fits my normal style.
    \item What's one personalized option that would likely suit my everyday interests?
    \item Give me one recommendation grounded in what I've consistently liked before.
\end{enumerate}

\paragraph{Recent-interest exploration requests.}
\begin{enumerate}
    \item I want to try something a bit new, but still something that feels like me. Any suggestion?
    \item Could you recommend one fresh option that connects to what I've been into lately?
    \item I'm open to exploring something new. What's one suggestion that still matches my taste?
    \item Choose one option that builds on what has caught my attention recently, without feeling random.
    \item I'd like a small change from my usual choices. What should I try?
    \item Recommend one new-ish activity or item that fits what I seem to be into right now.
    \item What's one recommendation that reflects what I've been paying attention to lately?
    \item I want something slightly outside my routine, but not totally unfamiliar. Any idea?
    \item Suggest one option that feels current for me while still matching my usual taste.
    \item Give me one practical recommendation that feels timely for me, not just my old favorites.
\end{enumerate}

\subsection{Dialogue Evaluation Judge Prompt}
\label{app:dialogue_judge_prompt}

\begin{promptbox}{Dialogue Evaluation Judge --- System Prompt}{judgepurple}
You are an expert judge for social-media-grounded personalized dialogue.

You will be given two independent recommendation cases, the user's gold profile, and the model responses.
The user requests are intentionally natural; do not require the model to mention post evidence or profile labels explicitly.

Important principles:
1. Reward semantic fit to the correct target interests, not exact wording similarity.
2. For stable_recommendation, reward use of stable interests.
3. For recent_interest_exploration, reward use of recent interests while keeping the suggestion compatible with stable interests.
4. Reward concrete, actionable, natural recommendations.
5. Penalize generic filler, unsupported assumptions, demographic guesses, and benchmark-like language.
6. The two cases are independent; do not require dialogue continuity across them.
7. Return strict JSON only.

Score each dimension from 0 to 5 using the rubrics below.
\end{promptbox}

\begin{promptbox}{Dialogue Evaluation Judge --- Rubrics}{ruleamber}
RUBRIC: interest_coverage
Whether the response engages the correct target interests (stable for stable_recommendation, recent for recent_interest_exploration) in this scenario.
- 0: No target interest is engaged; response is generic or irrelevant.
- 1: Tangential mention of a target interest but not used as the basis of the recommendation.
- 2: One target interest is partially engaged but the recommendation does not clearly center on it.
- 3: One target interest is clearly and centrally used as the basis of the recommendation.
- 4: Multiple target interests are used naturally, or one interest is used with specific supporting detail.
- 5: Rich, precise engagement with the correct target interests; the recommendation feels tailored to this specific user.

RUBRIC: concreteness
Whether the recommendation is specific, actionable, and natural rather than generic or vague.
- 0: Entirely generic; could apply to any user (e.g., "try something you enjoy").
- 1: A vague suggestion with no actionable detail.
- 2: Some concrete element is present but the recommendation remains broad.
- 3: A clear, actionable recommendation with at least one specific detail (e.g., a genre, activity, item, or place).
- 4: A specific recommendation with supporting context that makes it easy to act on.
- 5: A vivid, naturally-phrased recommendation with precise detail that feels like a human friend suggested it.

RUBRIC: fluency
Whether the response is well-formed, natural, and free of benchmark artifacts (e.g., scoring language, numbered lists, meta-commentary).
- 0: Incoherent or empty response.
- 1: Contains obvious benchmark artifacts (e.g., "turn 1:", "post index:", "score: 8/10", JSON remnants).
- 2: Awkward phrasing or overly structured language that does not read as natural dialogue.
- 3: Natural and readable but slightly stiff or formulaic.
- 4: Fluid and natural; reads like a real conversational assistant.
- 5: Fully natural, polished, and appropriate to the context; indistinguishable from human-written recommendation.
\end{promptbox}

\begin{promptbox}{Dialogue Evaluation Judge --- User Prompt}{judgepurple}
Evaluate this personalized dialogue prediction.

Use the scenario rubrics inside gold_reference_dialogue.
Score interest_coverage, concreteness, and fluency from 0 to 5.
Do not require explicit evidence explanation in the model response.

Do not copy the schema hint or return placeholder zeros. If all four scores are 0, brief_rationale must explain why.

{judge_input_json}

Return strict JSON only.
\end{promptbox}

\subsection{Caption Prompt}
\label{app:caption_prompt}

\begin{promptbox}{Image Captioning --- System Prompt}{systemblue}
You analyze one social-media image for downstream user-interest inference.

Return strict JSON only.

Rules:
- Focus on visible content only.
- Describe the main subject, activity, setting, mood/style, and any clearly readable text.
- Do not infer identity, demographics, private traits, or motivation.
- If the image is low-information, say that briefly.
- Keep summary concise and concrete.
\end{promptbox}

\subsection{Hierarchical Profile Construction Prompts}
\label{app:hierarchical_prompts}

\begin{promptbox}{Hierarchical --- Chunk Summarization --- System Prompt}{systemblue}
You are constructing an evidence-grounded user interest profile from a chronological
segment of a user's social-media timeline.

You will receive a sequence of posts. Each post may contain text, image captions, and a
timestamp. Your task is to summarize only the interests that are directly supported by
the posts in this segment.

Important rules:
1. Only infer interests that are supported by observable evidence in the posts.
2. Do not infer demographic attributes, personality traits, occupation, gender, age,
   race, religion, political identity, health status, or other sensitive personal
   attributes.
3. Distinguish recurring interests from one-off mentions.
4. Use both text and image captions as evidence.
5. Preserve post IDs as evidence anchors.
6. Do not over-generalize. For example, one photo of food does not mean the user is
   a food enthusiast unless there are repeated signals.
7. If the evidence is weak or incidental, mark it as weak.

Return valid JSON only.
\end{promptbox}

\begin{promptbox}{Hierarchical --- Global Aggregation --- System Prompt}{systemblue}
You are aggregating chunk-level summaries into a final user interest profile.

You will receive summaries from multiple chronological chunks of the same user's
social-media timeline. Your task is to merge redundant interests, identify stable and
recent interests, and produce a concise final profile.

Important rules:
1. Merge semantically equivalent interests. For example, "home cooking", "cooking
   meals", and "homemade food" should be normalized if they refer to the same core
   interest.
2. Stable interests should be supported across multiple posts or multiple time periods.
3. Recent interests should be supported by posts concentrated in the most recent part
   of the timeline, even if they are not stable.
4. Interests should only be included when supported by clear, repeated evidence; sparse
   or ambiguous signals should not be promoted to interests.
5. Do not infer sensitive attributes or demographics.
6. Do not create interests that are not supported by the provided chunk summaries.
7. Preserve evidence post IDs whenever possible.
8. Output valid JSON only.
\end{promptbox}

\subsection{Extractive Profile Construction Prompts}
\label{app:extractive_prompts}

The extractive--abstractive method proceeds in two stages. First, the LLM receives the full user timeline and is prompted to select up to $K$ representative posts per domain. The selection criteria include relevance, specificity (concrete interest signals rather than vague topics), recurrence (preferring posts consistent with repeated behavior), recency (capturing potential recent interests), and multimodal grounding (using image captions as evidence). Second, the LLM receives only the selected posts and synthesizes the final profile, without access to the original full timeline. The prompts for both stages are shown below.

\begin{promptbox}{Extractive --- Post Selection --- System Prompt}{systemblue}
You are selecting representative social-media posts for user interest profiling.

You will receive a user's chronological social-media timeline. Each post may include
text, image captions, and a timestamp. Your task is to select a small set of
representative posts for each interest domain. These selected posts will be used later
to generate the user's profile.

Important rules:
1. Select posts only when they provide concrete evidence for the domain.
2. Prefer posts that show recurring interests, strong visual/textual evidence, or
   recent concentrated activity.
3. Avoid selecting posts that only contain incidental, ambiguous, or very weak signals.
4. Use both text and image captions.
5. Do not infer sensitive attributes or demographics.
6. Do not summarize the profile yet. Only select representative posts.
7. Each domain can have at most the configured K selected posts.
8. If a domain has insufficient evidence, return an empty list for that domain.

Return valid JSON only.
\end{promptbox}

\begin{promptbox}{Extractive --- Abstractive Synthesis --- System Prompt}{systemblue}
You are generating an abstractive user interest profile from selected representative
social-media posts.

You will receive a small set of representative posts selected for each domain. Each
post may contain text, image captions, timestamp, and a selection reason. Your task is
to synthesize a concise, evidence-grounded user profile.

Important rules:
1. Use only the selected posts as evidence.
2. Do not infer interests that are not supported by selected posts.
3. Separate stable interests from recent interests.
4. Stable interests should be supported by multiple posts or recurring evidence.
5. Recent interests should be supported by posts concentrated in the most recent period.
6. Only include interests that are clearly supported; do not include interests when evidence is limited or ambiguous.
7. Do not infer sensitive attributes or demographic information.
8. Preserve supporting post IDs for every interest.
9. Output valid JSON only.
\end{promptbox}

\subsection{Calibration and Gold Rewrite Prompts}
\label{app:calibration_prompts}

\begin{promptbox}{Calibration --- Evidence-First Extraction --- System Prompt}{systemblue}
You are performing evidence-first interest summarization for one domain.

Goal:
- Produce natural-language interest labels and short descriptions suitable for
  downstream LLM personalization benchmarking.
- Use canonical tags only as evidence anchors, not as the main surface form.

Rules:
- Use only provided evidence.
- Output human-readable labels (2-6 words), not raw canonical ids.
- Labels should usually add user-facing detail beyond any single canonical tag.
  Synthesize repeated patterns from tag clusters, evidence examples, and
  representative posts.
- For each candidate interest, attach canonical_tags chosen only from the provided
  tag clusters.
- description must be 1 sentence of natural language.
- Exclude non-interest attributes such as family roles, career identity, and age range.
- Treat memes, reaction images, screenshots, and generic reposted aesthetic content
  as weak evidence by default.
- Do not infer a stable interest, ownership, or hobby from a single image when it
  could be a joke, repost, borrowed scene, or someone else's pet/car/food.
- For visual evidence, describe only what is directly visible. A photo of purchased
  food supports eating or dining evidence, not necessarily cooking.
- If evidence is sparse or fragmented, say so explicitly.
- Return strict JSON only.
\end{promptbox}

\begin{promptbox}{Calibration --- Domain LLM Calibration --- System Prompt}{systemblue}
You are calibrating a domain-level interest profile.

Your job:
1) Preserve natural-language candidate interests from pass1 whenever evidence
   supports them.
2) Use algorithmic statistics only to place interests into stable / recent / weak
   buckets.
3) Keep canonical_tags as anchors, but do not use raw canonical ids in labels or
   summaries.
4) Write domain_summary as 2-4 natural sentences suitable for downstream
   personalization benchmarking.

Hard constraints:
- Every interest item must contain both label (natural language) and canonical_tags
  (from provided clusters only).
- domain_summary should mention only labels that appear in structured fields.
- Exclude family roles, career identity, and age range from interest output.
- Do not treat memes, reaction images, screenshots, or generic reposts as strong
  evidence unless the user's personal engagement is repeated across posts.
- Do not infer pet ownership, cooking, driving, collecting, or other hands-on hobbies
  from a single ambiguous image.
- If no reliable interest can be extracted, explain that in plain language.
- Return strict JSON only.
\end{promptbox}

\begin{promptbox}{Calibration --- Benchmark Gold Rewrite --- System Prompt}{systemblue}
You are rewriting an internal domain analysis into a benchmark gold summary for user
profiling.

Goal:
- Convert an internal analysis-style summary into a natural-language profile summary
  suitable for evaluating LLM personalization ability.
- Keep the summary faithful to the analysis.
- Preserve caution, but reduce audit / pipeline wording.
- Make the result sound like a concise user-interest profile, not a system report.

Core requirements:
- Do NOT invent any new interests, hobbies, personality traits, demographics, or
  motivations.
- Treat stable_interests and recent_interests as the main factual anchors.
- Do not upgrade weak evidence into a firm preference.
- Do not turn inactive domains into meaningful preference descriptions.

Style requirements:
- Keep the tone concise, human-readable, and profile-oriented.
- Prefer 2-4 sentences.
- Avoid internal jargon such as support_posts, tag_clusters, weak_signal, stable_score.
- Prefer natural preference language such as "shows interest in", "tends to engage
  with", "shows limited signal around", "does not currently provide a reliable signal"
  when appropriate.

Output requirements:
- Return strict JSON only with schema: { "summary_natural_gold": string }
\end{promptbox}

\section{User Filtering and Profile Verification Details}
\label{app:user_filtering}

This appendix provides implementation details that are summarized in Section~\ref{subsec:user_selection} and Section~\ref{subsec:profile_construction}.

\paragraph{Automatic account filtering.}
We use three account-level heuristics before manual inspection. First, we retain accounts with 5--5{,}000 followers, excluding extremely inactive accounts and highly public accounts. Second, we compute the follower--followee ratio (FFR) as
\begin{equation}
    \mathrm{FFR} = \frac{N_{\mathrm{follower}}}{N_{\mathrm{followee}} + \epsilon},
\end{equation}
where $\epsilon$ avoids division by zero. We retain accounts with $\mathrm{FFR}\in[0.5,2]$, which favors relatively reciprocal social neighborhoods and filters highly asymmetric broadcaster-style accounts. Third, we compute the image trace density ratio (ITDR) as
\begin{equation}
    \mathrm{ITDR} = \frac{N_{\mathrm{posts\ with\ images}}}{N_{\mathrm{total\ posts}}},
\end{equation}
and retain users with $\mathrm{ITDR}\geq0.3$ to ensure sufficient visual evidence for multimodal profiling.

\paragraph{Manual account inspection.}
After automatic filtering, annotators remove accounts that are commercial, celebrity-like, organization-operated, repost-heavy, dominated by low-information content, or lacking sufficient personal and preference-relevant signals. This stage yields 250 candidate users for timeline collection and profile construction.

\paragraph{Cross-post aggregation.}
Post-level interest candidates are aggregated into canonical interests before temporal scoring. Near-duplicate posts are clustered and down-weighted so that repeated captions, repost-like content, or bursty discussions do not count as independent evidence. Semantically equivalent tags are normalized and merged into canonical interests. For each canonical interest, we retain supporting posts, modality attribution, duplicate-adjusted support, temporal distribution, and extraction confidence.

\paragraph{Human profile verification.}
Human verification is conducted over calibrated domain-level profiles and their supporting evidence. Annotators remove unsupported interests, revise overly broad labels into more concrete tags, and add missing tags when the evidence clearly supports them. Accepted labels must be concrete, non-sensitive, domain-appropriate, and supported by the timeline. After this verification stage, users with fewer than three active domains are removed to ensure enough positive personalization signals for evaluation. Our annotators are trained to be conservative and evidence-grounded, avoiding over-interpretation or inference beyond what the timeline supports. This process yields the final set of 100 users for the benchmark.

\section{Annotator Recruitment, Instructions, and Data Consent}
\label{app:ethics_checklist}

This section supplements the profile verification (Section~\ref{subsec:profile_construction}) and dialogue evaluation (Appendix~\ref{app:human_agreement}) with details on annotator recruitment, compensation, instructions, and data consent.

\paragraph{Recruitment and payment.}
All annotators were recruited from the undergraduate and graduate student population at the authors' institution. Annotators were compensated at 50 CNY per hour. This rate exceeds the typical student hourly wage at the authors' university and is consistent with compensation for comparable annotation tasks in the region.

\paragraph{Instructions to participants.}
All annotators received written annotation guidelines before beginning work, covering task definitions, quality rubrics, and annotated examples. For profile verification, annotators were instructed to be conservative and evidence-grounded, accepting interests only when directly supported by timeline evidence (see the criteria in Appendix~\ref{app:user_filtering}). For dialogue evaluation, annotators received the full 0--5 scoring rubrics shown in Appendix~\ref{app:dialogue_judge_prompt} and completed a calibration round before the agreement study. The complete instruction documents are available upon request.

\paragraph{Data consent.}
The social-media posts used in this benchmark were collected from publicly accessible accounts. We exclude private or restricted-access content. The released benchmark subset contains only de-identified text and captions; original images are excluded to prevent re-identification, and the full dataset is provided only through controlled research access (see Ethical Considerations). No annotator personal information was collected or retained.

\section{Profile Scoring Details}
\label{app:profile_scoring}

This appendix provides the detailed scoring rules used in the algorithmic temporal scoring stage of profile construction. These scores are used only to produce preliminary stable/recent assignments before LLM-based calibration and human verification.

\paragraph{Canonical interest evidence.}
After post-level extraction and cross-post aggregation, each canonical interest $c$ in domain $d$ is associated with an evidence set
\begin{equation}
    \mathcal{E}_{u,d,c} = \{ e_1, e_2, \dots, e_m \},
\end{equation}
where each evidence item corresponds to a supporting post. Each evidence item records the post timestamp, evidence modality, duplicate-adjusted weight, and post-level extraction confidence. If a post belongs to a near-duplicate cluster $C$, its support weight is discounted by $1/|C|$, so that repeated or highly similar posts do not artificially inflate an interest.

Let $w_j$ denote the duplicate-adjusted weight of evidence item $e_j$, and let $c_j \in [0,1]$ denote its extraction confidence. The effective support count of a canonical interest is defined as
\begin{equation}
    S_{\mathrm{eff}} = \sum_{e_j \in \mathcal{E}_{u,d,c}} w_j.
\end{equation}
The average confidence is computed as a weighted average:
\begin{equation}
    \bar{c}
    =
    \frac{\sum_{e_j \in \mathcal{E}_{u,d,c}} w_j c_j}
    {\sum_{e_j \in \mathcal{E}_{u,d,c}} w_j}.
\end{equation}

\paragraph{Temporal bins.}
To estimate whether an interest is persistent over time, we divide each user's timeline into monthly bins. Let $B_{\mathrm{distinct}}$ be the number of distinct monthly bins that contain at least one supporting evidence item for the canonical interest. Let $B_{\mathrm{span}}$ be the total number of monthly bins covered by the user's collected timeline. We set
\begin{equation}
    B_{\mathrm{req}} = \min(6, B_{\mathrm{span}}),
\end{equation}
so that long timelines require evidence spread across multiple periods, while shorter timelines are not penalized excessively.

\paragraph{Stable-interest score.}
The stable score estimates whether a canonical interest reflects a stable preference. It combines three factors: effective support, temporal dispersion, and extraction confidence:
\begin{equation}
\begin{aligned}
\mathrm{score}_{\mathrm{stable}}
&= 0.60 \min\!\left(1, \frac{S_{\mathrm{eff}}}{3}\right) \\
&\quad + 0.20 \bar{c}
   + 0.20 \min\!\left(1, \frac{B_{\mathrm{distinct}}}{m}\right),
\end{aligned}
\end{equation}
where $m = \max(2,B_{\mathrm{req}}-1)$.
The first term rewards repeated support after duplicate discounting. The second term incorporates the confidence of post-level extraction. The third term rewards evidence distributed across multiple time periods. This score is designed to favor interests that appear repeatedly and persistently across the user's timeline.

A canonical interest is marked as a preliminary stable interest if it satisfies
\begin{equation}
    \mathrm{score}_{\mathrm{stable}} \geq \theta_{\mathrm{stable}}
\end{equation}
and has sufficient temporal support:
\begin{equation}
    S_{\mathrm{eff}} \geq 3,
    \qquad
    B_{\mathrm{distinct}} \geq 2.
\end{equation}
In our implementation, we set
\begin{equation}
    \theta_{\mathrm{stable}} = 0.65.
\end{equation}

\paragraph{Recent-interest score.}
The recent score estimates whether a canonical interest reflects a recent or emerging preference. We focus on the most recent 90 days of the user's timeline. Let $\mathcal{E}_{u,d,c}^{90}$ denote the subset of evidence items whose timestamps fall within this window. We define
\begin{equation}
    S_{90} = \sum_{e_j \in \mathcal{E}_{u,d,c}^{90}} w_j,
\end{equation}
and
\begin{equation}
    \bar{c}_{90}
    =
    \frac{\sum_{e_j \in \mathcal{E}_{u,d,c}^{90}} w_j c_j}
    {\sum_{e_j \in \mathcal{E}_{u,d,c}^{90}} w_j}.
\end{equation}
If no evidence appears in the most recent 90 days, we set $S_{90}=0$ and $\bar{c}_{90}=0$.

The recent score is defined as
\begin{equation}
\mathrm{score}_{\mathrm{recent}}
=
0.65 \min\!\left(1, \frac{S_{90}}{2}\right)
+
0.35 \bar{c}_{90}.
\end{equation}
Compared with the stable score, the recent score places more emphasis on recent support and does not require broad temporal dispersion across the full timeline.

A canonical interest is marked as a preliminary recent interest if it does not satisfy the stable-interest condition but satisfies
\begin{equation}
    \mathrm{score}_{\mathrm{recent}} \geq \theta_{\mathrm{recent}}
\end{equation}
and has sufficient recent evidence:
\begin{equation}
    S_{90} \geq 2.
\end{equation}
In our implementation, we set
\begin{equation}
    \theta_{\mathrm{recent}} = 0.60.
\end{equation}

\section{Prompt Analysis: Conservative vs.\ Neutral Evaluation Instructions}
\label{app:prompt_analysis}

The profile evaluation prompt (Appendix~\ref{app:profile_eval_model_prompt}) instructs models to ``prefer fewer, broader tags'' and limit most active domains to ``1--2 reliable tags.'' This conservative design intentionally prioritizes precision over recall: without such guidance, models in pilot experiments produced noisy tag lists containing near-duplicates and one-off mentions that did not reflect genuine user interests.

We note a potential concern: if the prompt constrains output volume, the benchmark may measure prompt compliance rather than profiling capability, creating an artificial ceiling on recall. To rule this out, we conduct a prompt analysis replacing the conservative instructions with a \textbf{neutral} variant that removes all constraints on tag count and breadth. The full neutral system prompt is:

\begin{promptbox}{Profile Evaluation --- Neutral System Prompt}{systemblue}
You are evaluating a user's domain-level interests from social-media posts.

Your job is to decide whether a domain is active, and if so, extract all clearly supported interest tags.

Task:
- Decide whether the domain status is active or inactive.
- If active, extract the supported interest tags.
- If inactive, abstain cleanly: no interest tags.

Definitions:
- long_term_interest_tags: recurring, reinforced, or stable interests supported across multiple posts or over time.
- short_term_interest_tags: newer, narrower, or more time-local interests that are supported but not yet stable.

Rules for active domains:
- Return all clearly supported interest tags without artificially limiting the count.
- It is valid to return zero long_term_interest_tags.
- It is valid to return zero short_term_interest_tags.
- Each tag must be a short natural-language label that reflects a user interest theme, not a raw keyword list, hashtag list, named entity list, or one-off event.
- Prefer precise, specific tags that accurately reflect the user's demonstrated interests.
- Do not produce near-duplicate tags across long-term and short-term buckets.
- Do not split one core interest into both long-term and short-term tags unless the short-term tag adds a clearly distinct recent focus.
- Do not introduce unsupported themes, motivations, personality traits, or lifestyle claims.

Rules for inactive domains:
- long_term_interest_tags must be [].
- short_term_interest_tags must be [].

General rules:
- If the evidence is weak, sparse, one-off, or not clearly attributable to user preference, prefer inactive.
- Return strict JSON only.
\end{promptbox}

The neutral user prompt mirrors the same changes:

\begin{promptbox}{Profile Evaluation --- Neutral User Prompt}{usergreen}
Evaluate user profile signal for one domain.

DOMAIN: {domain}
DOMAIN_DEFINITION: {domain_definition}
OUTPUT_STATUS_OPTIONS: active or inactive

POSTS_CONTEXT:
{posts_context}

Instructions:
- Read the posts and decide whether this domain contains a reliable user-interest signal.
- If active, extract all clearly supported interest tags for this domain.
- Use long_term_interest_tags for recurring or stable interests.
- Use short_term_interest_tags for newer or more time-local interests.
- Return all clearly supported tags. Do not artificially limit the count.
- It is valid to return zero long_term_interest_tags.
- It is valid to return zero short_term_interest_tags.
- Use short natural-language labels that describe user interest themes rather than raw hashtags, named entities, or one-off events.
- Prefer precise, specific tags that accurately reflect the user's demonstrated interests.
- Do not produce near-duplicate tags across long-term and short-term buckets.
- Do not split one core interest into both long-term and short-term tags unless the short-term tag adds a clearly distinct recent focus.
- If the evidence is weak, sparse, one-off, or not clearly attributable to user preference, prefer inactive.
- If inactive, return empty tag lists.
- Return JSON only.
\end{promptbox}

We evaluate Qwen3.5-35B-A3B under the direct profile construction setting on the full 100-user evaluation subset, using identical gold profiles, anchor-matching, and evaluation protocol across both prompt variants.

\begin{table}[h]
\centering
\footnotesize 
\setlength{\tabcolsep}{6pt} 
\renewcommand{\arraystretch}{1.15}
\begin{tabular}{lccc}
\toprule
\textbf{Metric} & \textbf{Conservative} & \textbf{Neutral} & \textbf{$\Delta$} \\
\midrule
Interest Precision & 0.490 & 0.417 & $-$0.073 \\
Interest Recall    & 0.157 & 0.180 & +0.023 \\
Interest F1        & 0.238 & 0.252 & +0.014 \\
Active F1          & 0.788 & 0.812 & +0.025 \\
\bottomrule
\end{tabular}
\caption{Prompt analysis comparing a \textbf{Conservative} evaluation prompt (prefers fewer, broader tags) against a \textbf{Neutral} variant (no tag-count constraints). $\Delta = \text{Neutral} - \text{Conservative}$; positive values favor the neutral prompt. Results on Qwen3.5-35B-A3B under the direct setting with 100 users.}
\label{tab:prompt_analysis}
\end{table}

Removing the conservative constraints changes metrics only marginally. Interest Recall improves slightly (+0.023), but Interest Precision drops by more (--0.073) as models produce specific tags that misalign with gold labels. The net effect on Interest F1 is negligible (+0.014). These results confirm that the conservative prompt is not the primary driver of low recall---the gap reflects genuine model limitations in recovering fine-grained interests from behavioral traces, not an artifact of benchmark design.

\section{Input-Modality Analysis}
\label{app:input_ablation}

To quantify the relative contribution of each input modality to profiling performance, we evaluate two models under progressively richer input configurations: text only, image captions only, text plus image captions, and the full setting adding timestamps. Table~\ref{tab:input_ablation} reports active-domain detection and interest-tag recovery across all four conditions. Text provides the primary profiling signal, while image captions and timestamps contribute complementary gains, as discussed in Section~\ref{subsec:additional_analyses}.

\begin{table*}[t]
\centering
\footnotesize
\setlength{\tabcolsep}{6pt}
\renewcommand{\arraystretch}{1.15}
\begin{tabular}{l *{4}{c}}
\toprule
\textbf{Input mode} & \textbf{Active F1} & \textbf{Tag F1} & \textbf{Stable F1} & \textbf{Recent F1} \\
\midrule

\rowcolor{tabgray} \multicolumn{5}{c}{\textbf{Qwen3.5-35B-A3B}} \\
\midrule
Text only               & 0.6869        & 0.2198        & 0.2415         & 0.0088 \\
Image captions only     & 0.6724        & 0.2352        & 0.2782         & 0.0093 \\
Text + image captions   & 0.7692        & \best{0.2682} & 0.2846         & \best{0.0171} \\
Text + img. caps. + timestamp & \best{0.7875} & 0.2669        & \best{0.2989}  & 0.0133 \\
\midrule

\rowcolor{tabgray} \multicolumn{5}{c}{\textbf{GPT-4o-mini}} \\
\midrule
Text only               & 0.5033        & 0.1712        & 0.0745         & 0.0592 \\
Image captions only     & 0.4992        & 0.1527        & 0.1122         & 0.0352 \\
Text + image captions   & 0.7367        & \best{0.2796} & 0.1705         & 0.0812 \\
Text + img. caps. + timestamp & \best{0.7454} & 0.2785        & \best{0.1832}  & \best{0.0858} \\
\bottomrule
\end{tabular}
\caption{Input-modality analysis on the fixed 100-user evaluation subset. Timestamps are removed in the first three ablation settings for each model. The \textit{Text + img. caps. + timestamp} row corresponds to the direct profile construction setting from Table~\ref{tab:profiling_results} and serves as the full-input reference. \textbf{Stable F1} and \textbf{Recent F1} measure interest-tag recovery within the stable and recent temporal buckets respectively, computed via within-bucket optimal bipartite matching with cached LLM anchor judgments. Best values per column within each model group are \best{bolded}.}
\label{tab:input_ablation}
\end{table*}

\section{Case Studies: Error Analysis}
\label{app:case_studies}

This appendix presents detailed case studies supporting the error analysis in Sections~\ref{subsec:profiling_results} and~\ref{subsec:dialogue_results}. We examine predictions from representative users across all seven domains, comparing gold profiles against model outputs under the direct, hierarchical, and extractive settings. Tables~\ref{tab:case_overgeneralization}--\ref{tab:case_temporal} organize examples by failure pattern.

\medskip\noindent\textbf{Notation.} In all case-study tables that follow: S~=~stable interest tags; R~=~recent interest tags; D~=~direct, H~=~hierarchical, E~=~extractive--abstractive profiling.

\begin{table*}[t]
\centering
\footnotesize
\setlength{\tabcolsep}{5pt}
\renewcommand{\arraystretch}{1.20}
\begin{tabularx}{\textwidth}{@{}lclYc@{}}
\toprule
\rowcolor{tabgray}
\textbf{User} & \textbf{Posts} & \textbf{Active} & \textbf{Gold profile summary (stable $|$ recent interests, with evidence modality)} & \textbf{Inactive} \\
\midrule
User A & 115 & 5/7
& \textbf{sports\_outdoor}: outdoor recreation, hiking~[t+v] $|$ walking, park visit~[t+v]\par
  \textbf{entertainment}: music listening, book, creative writing~[t+v] $|$ poetry, reading, writing~[t+v]\par
  \textbf{food\_drink}: dining out, birthday cake, dessert, coffee~[t+v] $|$ restaurant dining, casual dining, confectionery~[t+v]\par
  \textbf{travel\_city}: sightsee, road trip, historic site, landmark~[t+v] $|$ Virginia Beach, roadside attraction, historical landmark~[t+v]\par
  \textbf{photography}: craft, photo sharing~[t+v] $|$ vintage photography~[t+v]
& gaming, pets \\
\midrule
User B & 100 & 6/7
& \textbf{sports\_outdoor}: football fandom~[t+v] $|$ soccer, Africa Cup, fitness~[t+v]\par
  \textbf{entertainment}: music listen, Arabic music~[t+v] $|$ Marwan Moussa, Spotify, movy~[t+v]\par
  \textbf{gaming}: eFootball~[t+v] $|$ video games, mobile gaming~[t+v]\par
  \textbf{food\_drink}: ---~$|$ iftar, breakfast, meal~[t+v]\par
  \textbf{travel\_city}: city exploration~[t+v] $|$ city walks, urban exploration~[t+v]\par
  \textbf{pets}: ---~$|$ cat~[v]
& photography \\
\midrule
User C & 187 & 5/7
& \textbf{sports\_outdoor}: wrestl, indie wrestling~[t+v]\par
  \textbf{entertainment}: wrestling, AEW, live event~[t+v] $|$ wrestle kingdom~[t+v]\par
  \textbf{food\_drink}: cocktail, home cooking~[t+v]\par
  \textbf{travel\_city}: sightsee, city exploration~[t+v]\par
  \textbf{photography}: event photography, photo editing~[t+v]
& gaming, pets \\
\midrule
User D & 185 & 6/7
& \textbf{sports\_outdoor}: walk, snorkeling, birdwatching~[t+v] $|$ nature observation, outdoor recreation~[t+v]\par
  \textbf{entertainment}: classic rock, the grinch, music listening~[t+v] $|$ concert attendance, music~[t+v]\par
  \textbf{food\_drink}: pub, pub visit, coffee, breakfast~[t+v] $|$ banana bread~[t+v]\par
  \textbf{travel\_city}: sightsee, city exploration, cruise travel, city walks~[t+v] $|$ Volendam, Netherlands, Sinai desert~[t+v]\par
  \textbf{photography}: landscape photography, photography, nature photography, night photography~[t+v] $|$ bird photography, black and white photography, outdoor photography~[t+v]\par
  \textbf{pets}: dog walk, dog ownership, dog care, pet ownership~[t+v] $|$ pet friendly pub~[t+v]
& gaming \\
\bottomrule
\end{tabularx}
\caption{Overview of the four case-study users. For each user we report the number of posts in the observation window, the count of active domains out of seven, a compact gold profile summary with evidence modality annotations (t~=~text, v~=~visual), and the inactive domains. A dash (---) in the stable or recent slot means no interests of that type were annotated for the domain. Detailed per-domain breakdowns appear in the individual case-study tables below.}
\label{tab:user_snapshots}
\end{table*}

\subsection{Over-Generalization and Cross-Category Confusion}

Table~\ref{tab:case_overgeneralization} illustrates the most pervasive failure mode: models collapsing multiple specific gold tags into one or two broad categories, and confusing semantically adjacent but factually incorrect categories.

\begin{table*}[t]
\centering
\footnotesize
\setlength{\tabcolsep}{6pt}
\renewcommand{\arraystretch}{1.15}
\begin{tabularx}{\textwidth}{@{}L{0.09\textwidth}L{0.14\textwidth}L{0.32\textwidth}Y@{}}
\toprule
\rowcolor{tabgray}
\textbf{User} & \textbf{Domain} & \textbf{Gold tags} & \textbf{Model predictions} \\
\midrule
User A & food\_drink
& \textbf{S}: dining out, birthday cake, dessert, coffee\par
  \textbf{R}: restaurant dining, casual dining, confectionery
& Gemini-2.5 (D): S=\{Dining out, \textbf{Baking}\}\par
  GPT-5.4 (D): S=\{sweets and desserts, restaurants and dining out\}\par
  GPT-4o-mini (D): \textbf{INACTIVE}\par
  Qwen3-VL-8B (D): \textbf{INACTIVE} \\
\midrule
User B & entertainment
& \textbf{S}: music listen, Arabic music\par
  \textbf{R}: Marwan Moussa, Spotify, movy
& Gemini-2.5 (D): S=\{music, movies and TV\}\par
  GPT-5.4 (D): S=\{Arabic music\} \\
\bottomrule
\end{tabularx}
\caption{Over-generalization and cross-category confusion. Models reduce 4--7 gold tags to 1--2 broad labels, hallucinate factually wrong categories (e.g., ``Baking'' for a user who only dines out and buys desserts), or predict INACTIVE for domains with abundant evidence. \texttt{(D)}~=~direct setting.}
\label{tab:case_overgeneralization}
\end{table*}

\begin{postbox}{Representative posts: User A food\_drink evidence (dining out and store-bought desserts, no home cooking)}
\footnotesize
\RaggedRight
\setlength{\parindent}{0pt}
\setlength{\parskip}{0pt}
\emergencystretch=1.5em

\textpostentry{2026-01-01}{New Year dining out}{Last night, for \#NewYear2026, we went out to eat. As I was biting my delicious chicken strip, I thought about all who couldn't be out, due to being sick, bedridden with cancer, frail unable to walk.}

\postentry{2026-01-13}{store-bought birthday cake}{Hubs birthday soon. My mom always bought Pepperidge Farms cakes for birthday celebrations. Sure miss her.}{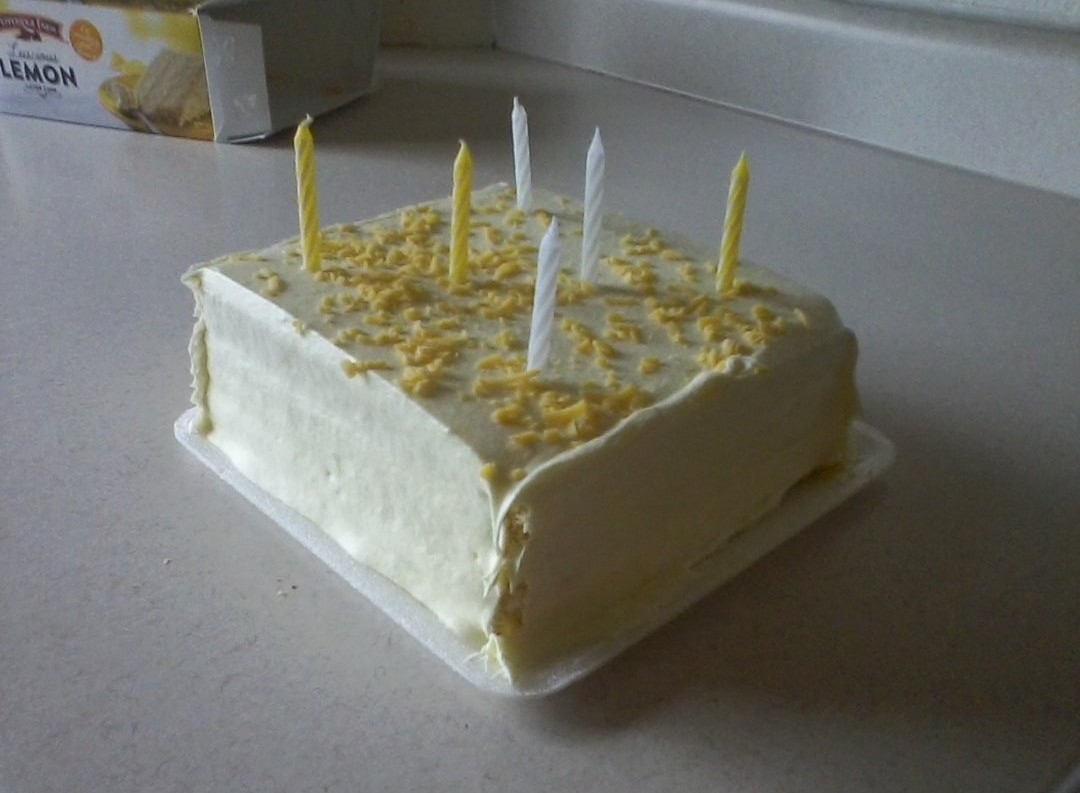}{34mm}

\postentry{2026-01-05}{restaurant dining}{I miss eating Paradiso with my mom and hubs together.}{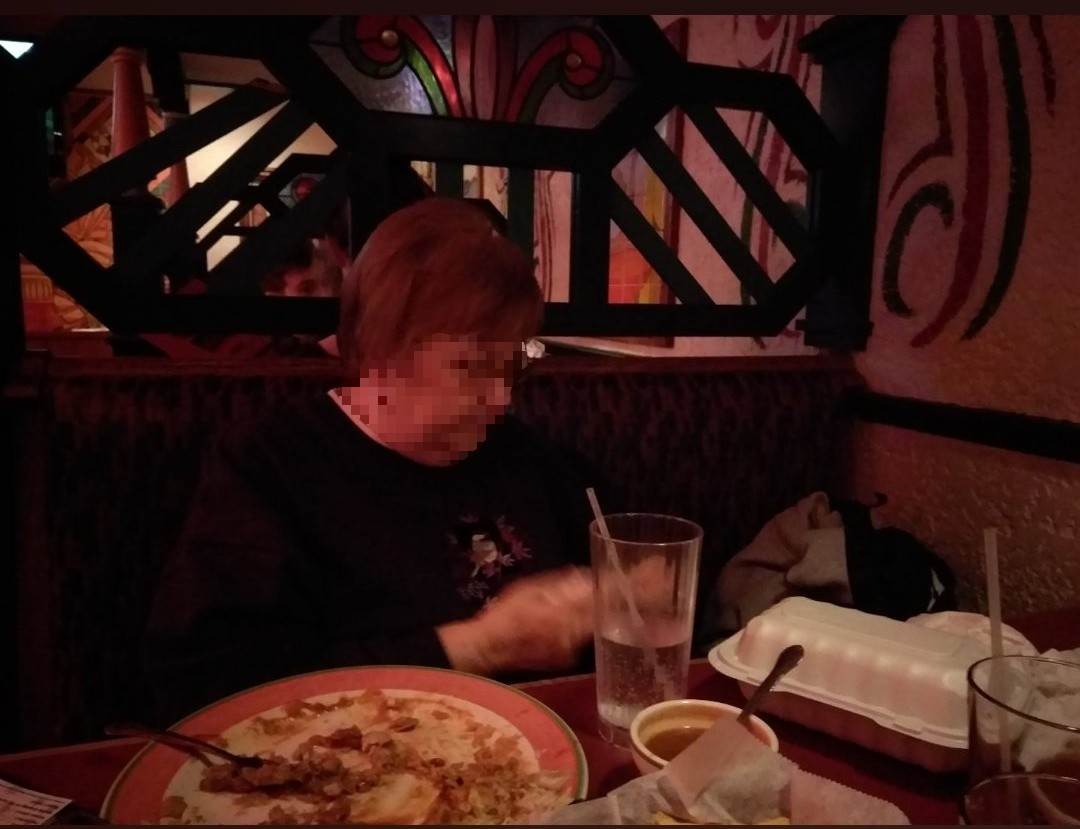}{34mm}

\vspace{3pt}
{\scriptsize\itshape
These three posts illustrate the user's food\_drink profile: dining out on New Year's Eve, purchasing store-bought Pepperidge Farms cakes for a birthday, and eating at a restaurant (Paradiso). The user's gold profile includes \textbf{home cooking} as a \emph{negative} interest---this user does not cook at home. Yet Gemini-2.5 predicts ``Baking'' with no evidence of any baking activity anywhere in the timeline. GPT-5.4 over-generalizes seven specific tags into two broad categories. GPT-4o-mini and Qwen3-VL-8B predict INACTIVE despite seven gold tags supported by posts spanning the full four-month window.
\par}
\end{postbox}

The User~A food\_drink case exhibits three distinct but related failure modes in a single domain. First, \textbf{over-generalization}: GPT-5.4 collapses seven specific gold tags---\texttt{dining out}, \texttt{birthday cake}, \texttt{dessert}, \texttt{coffee}, \texttt{restaurant dining}, \texttt{casual dining}, and \texttt{confectionery}---into just two broad labels (\texttt{sweets and desserts}, \texttt{restaurants and dining out}). While these are not factually wrong, they discard the granularity needed for downstream personalization: recommending a confectionery shop is qualitatively different from recommending a birthday cake bakery, and a coffee shop recommendation differs from a casual-dining suggestion. Second, \textbf{cross-category confusion}: Gemini-2.5 predicts \texttt{Baking} despite the user having zero baking activity anywhere in the timeline. The gold profile explicitly includes \texttt{home cooking} as a \emph{negative} interest---this user purchases prepared food and eats at restaurants, but does not cook. The model appears to conflate ``engages with food content'' with ``prepares food at home,'' a category error analogous to confusing ``attends concerts'' with ``plays an instrument.'' Third, \textbf{false inactive}: GPT-4o-mini and Qwen3-VL-8B classify the entire domain as INACTIVE, missing all seven gold tags. This is a severe false-inactive error: the domain is supported by posts spanning the full four-month observation window across multiple modalities (text, images, and mixed), yet two models fail to activate it at all.

The User~B entertainment case exhibits the same \textbf{over-generalization} pattern. Gemini-2.5 predicts \texttt{music, movies and TV}, adding a film/television interest absent from the gold profile, while GPT-5.4 reduces five tags to one (\texttt{Arabic music}). Across both cases, the pattern is consistent: models default to broad, safe hypernyms and resist committing to the specific subcategories that make personalized recommendations actionable.

\subsection{Domain Blindness: False Inactive Errors}

Table~\ref{tab:case_false_inactive} shows cases where models incorrectly classify an active domain as inactive, missing all gold interest tags. These errors concentrate in domains whose evidence is primarily textual and dispersed across many posts.

\begin{table*}[t]
\centering
\footnotesize
\setlength{\tabcolsep}{6pt}
\renewcommand{\arraystretch}{1.15}
\begin{tabularx}{\textwidth}{@{}L{0.09\textwidth}L{0.14\textwidth}L{0.30\textwidth}Y@{}}
\toprule
\rowcolor{tabgray}
\textbf{User} & \textbf{Domain} & \textbf{Gold tags} & \textbf{Which models missed it} \\
\midrule
User B & travel\_city & S: city exploration\par R: city walks, urban exploration & All four models (D) \\
\midrule
User B & food\_drink & R: iftar, breakfast, meal & GPT-5.4, Qwen3.5, Qwen2.5 (D) \\
\bottomrule
\end{tabularx}
\caption{False inactive errors. Domains supported primarily by scattered textual mentions are frequently missed entirely, even by strong models. \texttt{(D)}~=~direct setting.}
\label{tab:case_false_inactive}
\end{table*}

\begin{postbox}{Representative posts: User B city-exploration evidence (text-dispersed domain)}
\footnotesize
\RaggedRight
\setlength{\parindent}{0pt}
\setlength{\parskip}{0pt}
\emergencystretch=1.5em

\textpostentry{2026-01-07}{text only, no image}{I wanna migrate illegally and the guys are like let's go to Ifrane hhhhhhhhhhhhh}

\postentry{2026-02-19}{city walk (4 images)}{Late night walk}{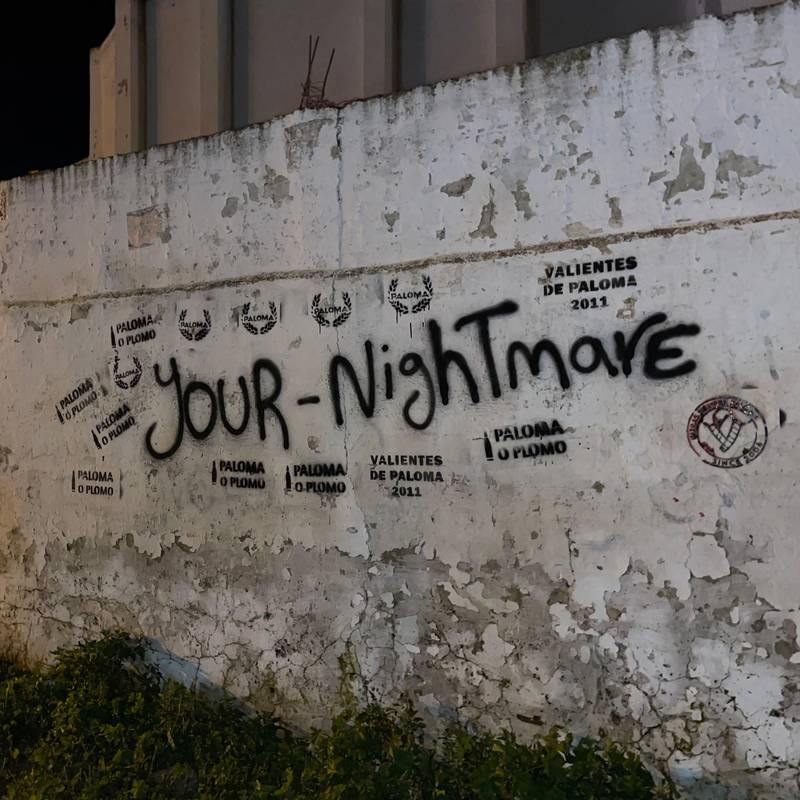}{34mm}

\postentry{2026-02-23}{urban exploration (1 image)}{Bars}{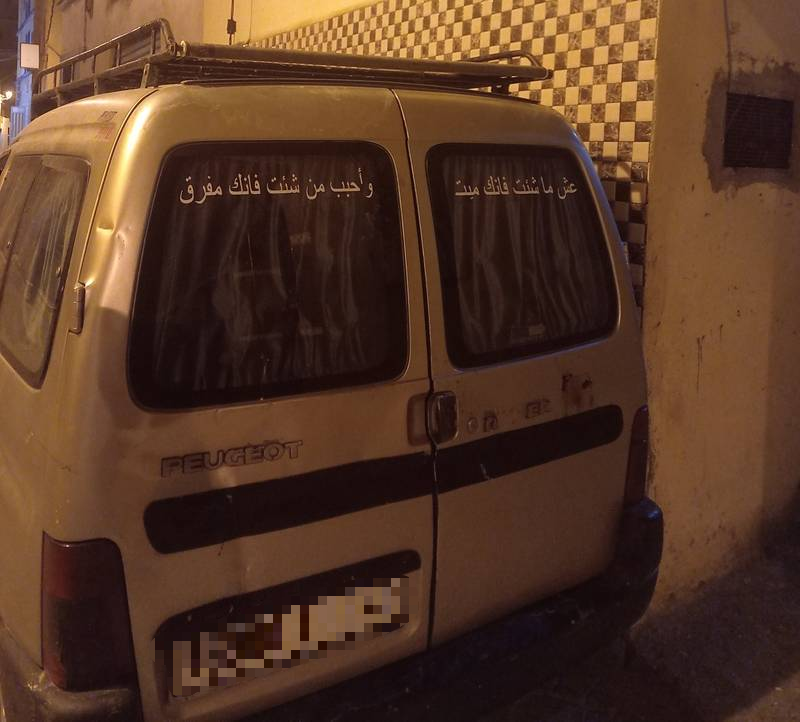}{34mm}

\vspace{3pt}
{\scriptsize\itshape
These three posts collectively support \textbf{city exploration} (stable) and \textbf{city walks} / \textbf{urban exploration} (recent). However, the evidence is distributed across short, casual text mentions with no hashtags, no location tags in the text, and no visually distinctive landmarks. All four models classified this domain as inactive when using the direct setting, because without a visually anchoring post, the weak textual signal fails to reach the activation threshold.
\par}
\end{postbox}

The User B travel case is illustrative: all four models predict inactive for a domain where gold lists \texttt{city exploration}, \texttt{city walks}, and \texttt{urban exploration}. These interests are expressed through casual text mentions across posts (e.g., ``went for a walk downtown,'' ``exploring a new neighborhood'') rather than through prominent images or hashtags. Without a visually anchoring post, the evidence fails to reach the model's activation threshold.

\subsection{Hallucinated Domains: False Active Errors}

The inverse failure---activating a domain the user does not actually engage with---occurs when models over-interpret incidental posts as preference signals. Table~\ref{tab:case_false_active} presents two representative cases. The first involves a user whose timeline is dominated by professional-wrestling content (187 posts, gold-active domains: sports\_outdoor and entertainment with wrestling-related interests). GPT-5.4 under hierarchical profiling activates the gaming domain with tags that explicitly name wrestling---``video game references in wrestling-related memes'' and ``Pokémon-themed wrestling events.'' The model's own labels concede the content is wrestling-related, yet it places them in gaming. Five posts out of 187 mention video games at all, and all five are wrestling-context posts: three document a CMLL$\times$Pokémon crossover wrestling show, one jokes about a wrestler taking time off to play a game, and one is an incidental mention. The model conflates \emph{wrestling content that references gaming} with \emph{the user having a gaming interest}.

\begin{table*}[t]
\centering
\footnotesize
\setlength{\tabcolsep}{4pt}
\renewcommand{\arraystretch}{1.22}
\begin{tabularx}{\textwidth}{@{}L{0.10\textwidth}L{0.15\textwidth}L{0.20\textwidth}Y@{}}
\toprule
\rowcolor{tabgray}
\textbf{User} & \textbf{Domain} & \textbf{Gold status} & \textbf{Model predictions} \\
\midrule
User C & gaming & inactive
& GPT-5.4 (H): R=\{\,video game references in wrestling-related memes,\; Pokémon-themed wrestling events\,\}\par
  GPT-5.4 (E): R=\{\,video game releases and references\,\} \\
\midrule
User B & photography & inactive
& GPT-5.4 (D): S=\{\,selfie and portrait photography\,\}\par
  GPT-5.4 (H): R=\{\,selfie portraits and self-image/profile visual curation\,\}\par
  Qwen3-VL-8B (H): R=\{\,image editing and AI-generated visuals\,\} \\
\bottomrule
\end{tabularx}
\caption{False active errors: category confusion and systemic over-interpretation of incidental post content. D~=~direct, H~=~hierarchical, E~=~extractive--abstractive.}
\label{tab:case_false_active}
\end{table*}

\begin{postbox}{Representative posts: User C wrestling content that triggered gaming hallucination}
\footnotesize
\RaggedRight
\setlength{\parindent}{0pt}
\setlength{\parskip}{0pt}
\emergencystretch=1.5em

\textpostentry{2025-09-26}{CMLL$\times$Pokémon crossover wrestling show}{Checking out that \#CMLL Pokémon show for a bit. This already looks like so much fun and I can't believe they got approval from Nintendo for it. \#LeyendasPokémonZA}

\postentry{2025-09-26}{same event}{The commitment from this fan to rock the Umbreon gimp mask for the show \dots\ \#CMLL \#LeyendasPokémonZA}{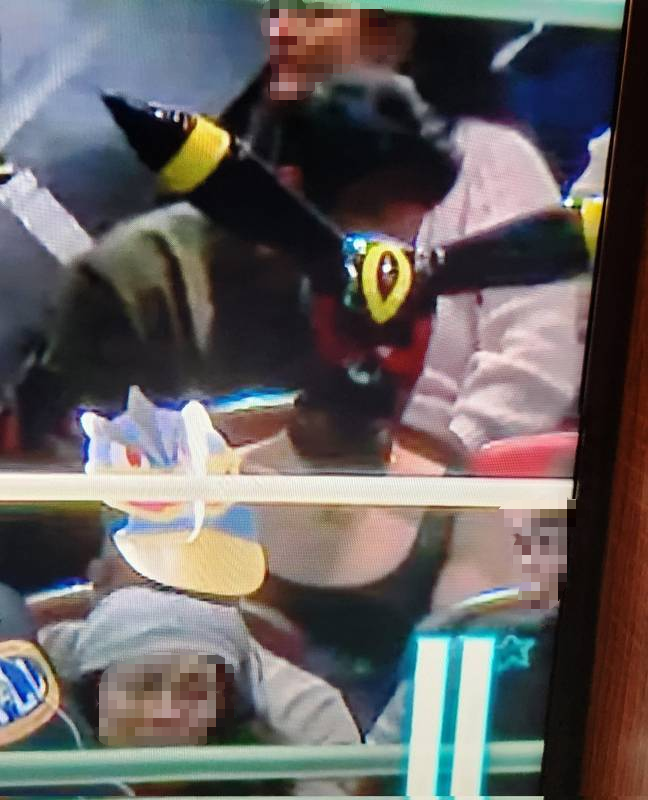}{34mm}

\textpostentry{2025-03-13}{wrestling joke referencing Assassin's Creed}{It's okay, Will. We know you need time off to play the new Assassin's Creed game coming out next week. Don't need to use your wife as an excuse. \#AEWDynamite}

\postentry{2025-10-19}{AEW stage design compared to Borderlands}{The St.\ Louis arch on the \#AEWWrestleDream stage is now making me think of the Borderlands vaults and all the wrestlers are just different Vault Hunters. \#AEW}{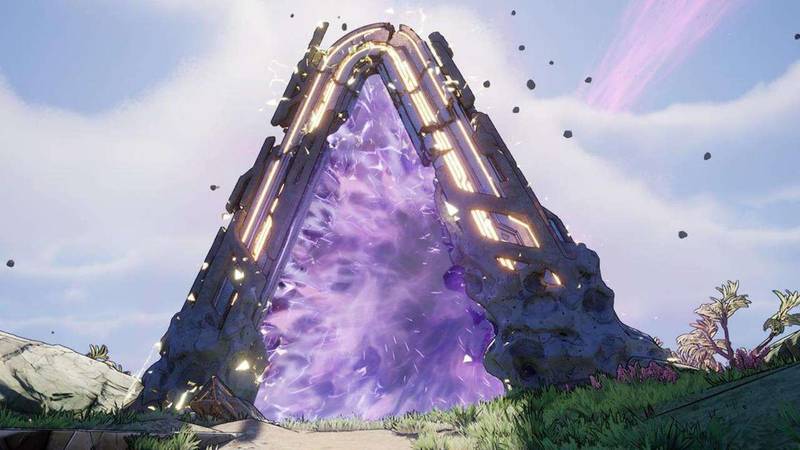}{34mm}

\postentry{2025-10-24}{Undertale meme, wrestling reaction}{``Hopes and Dreams'' intensifies. \#AEW \#Undertale}{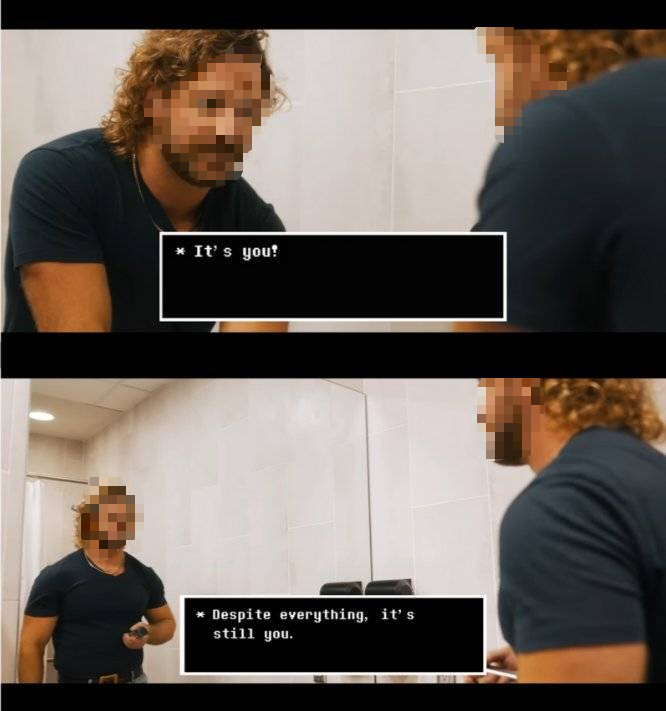}{34mm}

\vspace{3pt}
{\scriptsize\itshape
These five posts are the only video-game mentions across 187 posts. All are contextual to professional wrestling: three document a wrestling show with a Pokémon promotional crossover, one jokes about a wrestler taking time off, one compares an AEW stage design to Borderlands, and one uses an Undertale meme to react to a match. GPT-5.4's own predicted tags concede the content is ``wrestling-related''---yet the model activates the gaming domain rather than recognizing this as wrestling-fan content that belongs in the user's already-active sports\_outdoor and entertainment domains.
\par}
\end{postbox}

The User~B photography case reveals a broader systemic pattern: across the full benchmark, posting selfies---a common behavior on social media---is frequently misinterpreted as photography enthusiasm. This mirrors the over-generalization pattern from Section~\ref{app:case_studies}: models lack the pragmatic judgment to distinguish ``taking a photo to document an experience'' from ``photography as a sustained interest.'' The User~C case adds a further dimension: even when the model correctly identifies the \emph{topic} of a post (wrestling), it can assign it to the wrong domain, revealing a category-boundary problem in how models map post content to interest domains.

\subsection{Recent-Interest Blindness and Temporal Bucket Confusion}

Table~\ref{tab:case_temporal} illustrates a pervasive failure: models cannot distinguish long-standing interests from recently emerged ones, assigning nearly all predictions to the stable bucket regardless of when evidence appears in the timeline. User~D provides a representative case: a frequent traveler with a clear pattern of general, recurring travel behavior (cruises, city walks, sightseeing) punctuated by specific recent destinations (Volendam, the Netherlands, the Sinai Desert).

\begin{table*}[t]
\centering
\footnotesize
\setlength{\tabcolsep}{6pt}
\renewcommand{\arraystretch}{1.15}
\begin{tabularx}{\textwidth}{@{}L{0.09\textwidth}L{0.17\textwidth}L{0.28\textwidth}Y@{}}
\toprule
\rowcolor{tabgray}
\textbf{User} & \textbf{Domain} & \textbf{Gold temporal split} & \textbf{Predictions (stable $|$ recent)} \\
\midrule
User D & travel\_city
& \textbf{S}: 4 tags: sightsee, city exploration, cruise travel, city walks\par
  \textbf{R}: 3 tags: volendam, netherland, sinai desert
& GPT-5.4 (H): S=\{Caribbean cruise, city exploration, \textbf{Amsterdam}\}\par
  \hspace{1.6em}R=\{Egypt, Sinai, \textbf{Netherlands}\}\par
  Qwen3.5 (H): S=\{International Travel, Cruise Travel\}, R=0\par
  Qwen3-VL (E): S=\{cruise-based Caribbean exploration\}, R=0 \\
\bottomrule
\end{tabularx}
\caption{Recent-interest blindness and temporal bucket confusion. Models fail to distinguish stable from recent interests, assigning predictions indiscriminately to the stable bucket. H~=~hierarchical, E~=~extractive--abstractive.}
\label{tab:case_temporal}
\end{table*}

\begin{postbox}{Representative posts: User D travel evidence illustrating stable vs.\ recent temporal structure}
\footnotesize
\RaggedRight
\setlength{\parindent}{0pt}
\setlength{\parskip}{0pt}
\emergencystretch=1.5em

\postentry{2025-11-28}{stable: cruise travel}{Marella Discovery 2, our home for the next two weeks at Berth in Bridgetown Port, Barbados as we and a multitude of other new passengers wait to board.}{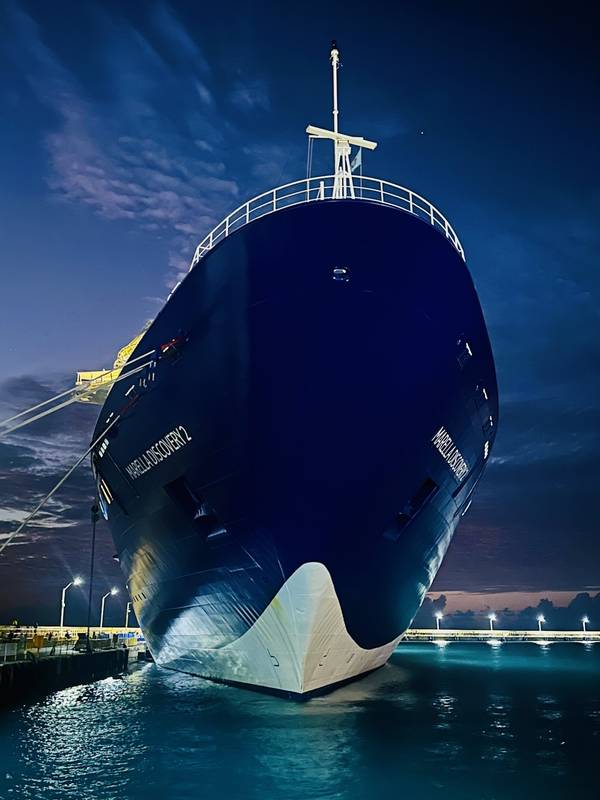}{34mm}

\postentry{2025-12-19}{stable: city walks}{Happy \#FingerpostFriday from \#Budapest\dots\ I came across these cyclist friendly fingerposts on a late evening walk by the River \#Danube.}{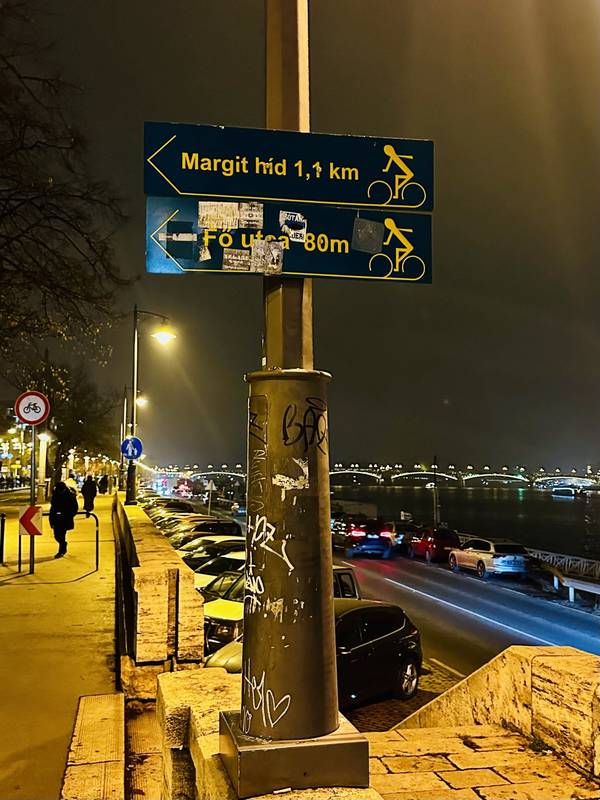}{34mm}

\postentry{2026-01-16}{recent: Netherlands/Volendam}{Happy \#FingerpostFriday\dots\ Here's a set of Fingerposts from the lovely town of \#Volendam in the Netherlands that we visited on a day trip out from Amsterdam.}{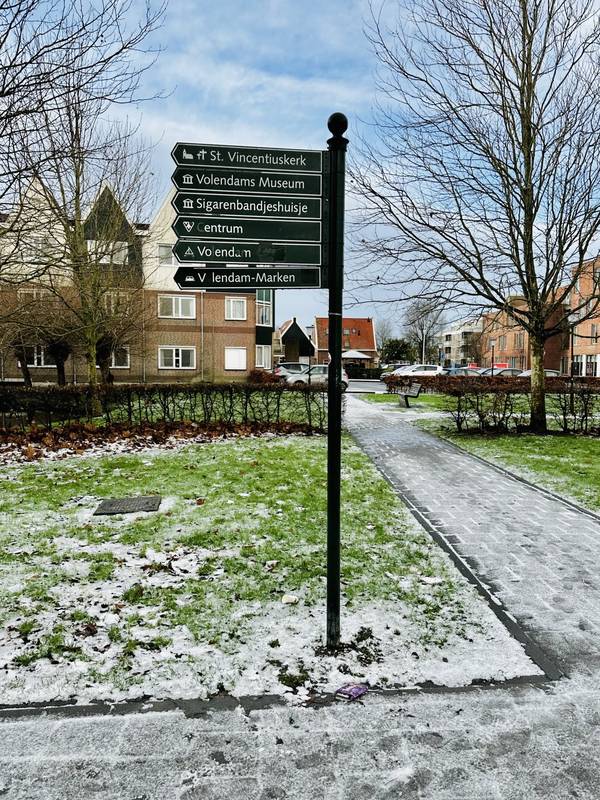}{34mm}

\postentry{2026-02-15}{recent: Sinai Desert}{Sunset Sinai Desert style. A fabulous excursion out into the desert this afternoon, to see a stark but stunning landscape that just seems utterly timeless.}{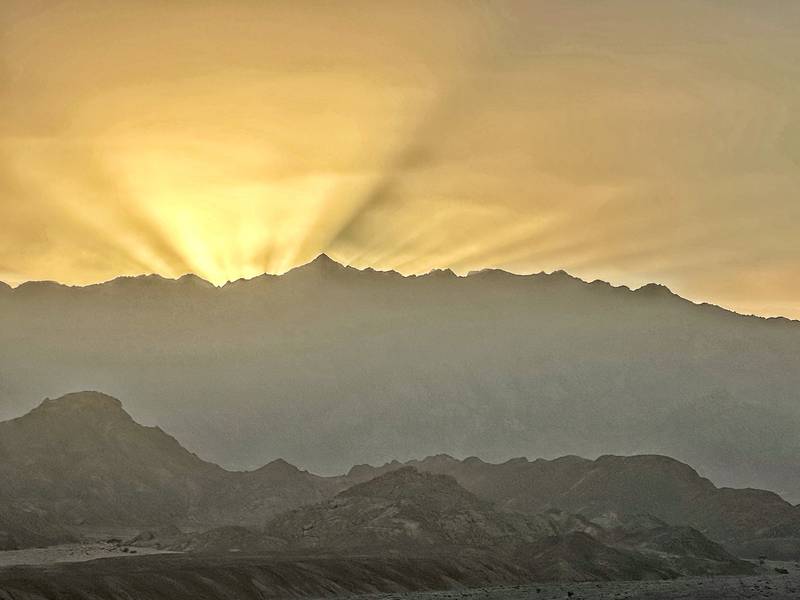}{34mm}

\vspace{3pt}
{\scriptsize\itshape
The temporal structure of User~D's travel domain is clear from the timeline: cruises, city walks, and sightseeing recur across the full four-month timeline (stable), while Volendam (January), the Netherlands (January--February), and the Sinai Desert (February) are specific destinations visited only in the final two months. GPT-5.4 partially captures this---placing Egypt and Netherlands travel in the recent bucket---but contradicts itself by assigning ``Amsterdam city exploration'' to the stable bucket, even though Amsterdam and the Netherlands are the same trip. Qwen3.5 and Qwen3-VL exhibit complete temporal blindness: they collapse all evidence into one or two broad stable tags and assign zero predictions to the recent bucket.
\par}
\end{postbox}

The failure pattern extends beyond User~D. Across the benchmark, models assign 70--90\% of predictions to the stable bucket regardless of temporal evidence distribution. When recent tags are predicted, they often correspond to interests that the gold profile classifies as stable, and vice versa. GPT-5.4's Amsterdam/Netherlands contradiction is especially revealing: the model recognizes that ``the Netherlands'' is a recent topic but places the capital city of that same country in the stable bucket, demonstrating that these models perform surface-level topic labeling rather than genuine temporal reasoning about when and how frequently evidence appears.

\subsection{Dialogue Case Studies}
\label{app:dialogue_cases}

The following examples illustrate how profiling failures propagate into downstream dialogue (Section~\ref{subsec:dialogue_results}).

\paragraph{Profile under-generation case (food-drink user).}
In the direct-profile-conditioned setting, a user's gold food-drink profile contains seven specific interests spanning dining out, regional cuisines, and holiday meals. Gemini-2.5-Flash's profile reduces this to two tags (\textit{Dining out}, \textit{Asian cuisine}); its dialogue response ignores food entirely, instead recommending astrophotography based on the user's photography interests. The judge assigns coverage~$=$~1.0/5, noting the model ``ignores the required interests.'' GPT-5.4's profile for the same user captures \textit{restaurant dining} and \textit{sushi}; its dialogue recommends a specific Japanese restaurant, earning coverage~$=$~3.0/5. The difference illustrates a compounding failure: conservative profiling removes the very tags that would enable diverse, personalized responses.

\paragraph{Visual dominance case (astrophotography fixation).}
In the timeline-conditioned setting, models can access all posts directly, yet they exhibit the same modality bias observed in profile construction. For the user described above, Gemini-2.5-Flash overlooks seven supported food interests and a hiking interest to recommend astrophotography across both dialogue turns, fixating on the domain with the most visually prominent evidence (sky and landscape photography). This pattern recurs across users: when a domain generates abundant images, it crowds out text-supported interests in downstream dialogue, even when those text-supported interests are equally or more relevant to the user's request.

These two cases illustrate a compounding failure chain. Conservative profiling strips away fine-grained interest tags (profile under-generation); the resulting sparse profile provides insufficient hooks for the dialogue model, which then defaults to visually dominant domains regardless of the user's actual request. The chain can be broken at either stage---better profiling yields richer profiles, and direct timeline access bypasses profile sparsity---but the modality bias persists in both paths, suggesting that balanced cross-modal attention is a prerequisite for either approach to succeed.

\section{Profile--Dialogue Correlation Tables}
\label{app:correlation_tables}

This appendix provides the full correlation tables referenced in Section~\ref{subsec:additional_analyses}.

\begin{table*}[t]
\centering
\footnotesize
\setlength{\tabcolsep}{6pt}
\renewcommand{\arraystretch}{1.15}
\begin{tabular}{l *{3}{c}}
\toprule
\textbf{Model} & \textbf{Interest F1 $\leftrightarrow$ Avg.} & \textbf{Interest F1 $\leftrightarrow$ Coverage} & \textbf{Interest F1 $\leftrightarrow$ Concreteness} \\
\midrule

\rowcolor{tabgray} \multicolumn{4}{c}{\textbf{Direct}} \\
\midrule
Gemini-2.5-Flash       & $-0.016$   & $+0.013$   & $+0.039$ \\
GPT-4o-mini            & $+0.221^{*}$ & $+0.245^{*}$  & $+0.189$ \\
GPT-5.4                & $+0.128$    & $+0.113$    & $+0.087$ \\
Qwen2.5-VL-7B-Instruct & $+0.094$    & $+0.299^{**}$  & $+0.319^{**}$ \\
Qwen3-VL-8B-Instruct   & $+0.092$    & $+0.240^{*}$  & $+0.224^{*}$ \\
Qwen3.5-35B-A3B        & $-0.042$    & $+0.089$    & $+0.054$ \\
\midrule

\rowcolor{tabgray} \multicolumn{4}{c}{\textbf{Hierarchical}} \\
\midrule
Gemini-2.5-Flash       & $+0.330^{***}$ & $+0.472^{***}$ & $+0.177$ \\
GPT-4o-mini            & $+0.148$  & $+0.319^{**}$ & $+0.028$ \\
GPT-5.4                & $+0.214^{*}$  & $+0.244^{*}$  & $+0.306^{**}$ \\
Qwen2.5-VL-7B-Instruct & $+0.102$    & $+0.477^{***}$ & $+0.256^{*}$ \\
Qwen3-VL-8B-Instruct   & $+0.065$    & $+0.251^{*}$    & $+0.157$ \\
Qwen3.5-35B-A3B        & $+0.207^{*}$    & $+0.311^{**}$ & $+0.233^{*}$ \\
\midrule

\rowcolor{tabgray} \multicolumn{4}{c}{\textbf{Extractive}} \\
\midrule
Gemini-2.5-Flash       & $+0.208^{*}$    & $+0.484^{***}$ & $+0.153$ \\
GPT-4o-mini            & $+0.198^{*}$    & $+0.222^{*}$    & $+0.055$ \\
GPT-5.4                & $+0.172$  & $+0.204^{*}$  & $+0.128$ \\
Qwen2.5-VL-7B-Instruct & $+0.138$    & $+0.446^{***}$ & $+0.245^{*}$ \\
Qwen3-VL-8B-Instruct   & $+0.022$    & $+0.413^{***}$ & $+0.722^{***}$ \\
Qwen3.5-35B-A3B        & $+0.186$  & $+0.318^{**}$ & $+0.193$ \\
\bottomrule
\end{tabular}
\caption{Per-user Spearman rank correlation between profile Interest Tag F1 and dialogue quality dimensions across three profile-conditioned settings. Each cell reports $\rho$ over 89--93 users (model-dependent, after excluding generation or judge errors). Significance: $^{*}p<0.05$, $^{**}p<0.01$, $^{***}p<0.001$.}
\label{tab:correlation_results}
\end{table*}

\begin{table*}[t]
\centering
\footnotesize
\setlength{\tabcolsep}{3pt}
\renewcommand{\arraystretch}{1.15}
{%
\renewcommand{\tabularxcolumn}[1]{m{#1}}%
\begin{tabularx}{\linewidth}{@{}m{0.22\linewidth}Y*{4}{>{\centering\arraybackslash}m{0.135\linewidth}}@{}}
\toprule
\textbf{Model} & \textbf{Stable F1 $\leftrightarrow$ Stable Rec.} & \textbf{Recent F1 $\leftrightarrow$ Recent Expl.} & \textbf{Stable F1 $\leftrightarrow$ Recent Expl.} & \textbf{Recent F1 $\leftrightarrow$ Stable Rec.} \\
\midrule
Gemini-2.5-Flash       & $+0.415^{***}$ & $+0.189$ & $+0.154$ & $+0.336^{***}$ \\
GPT-4o-mini            & $+0.172$       & $+0.265^{*}$  & $+0.254^{*}$  & $+0.048$ \\
GPT-5.4                & $+0.038$  & $+0.186$ & $-0.060$ & $+0.317^{**}$ \\
Qwen2.5-VL-7B-Instruct & $+0.496^{***}$ & $+0.385^{**}$  & $+0.364^{**}$  & $+0.466^{***}$ \\
Qwen3-VL-8B-Instruct   & $-0.067$ & $-0.109$ & $-0.009$ & $+0.217^{*}$ \\
Qwen3.5-35B-A3B        & $+0.214^{*}$       & $+0.161$ & $-0.018$ & $-0.036$ \\
\bottomrule
\end{tabularx}%
}
\caption{Temporal breakdown of profile--dialogue correlation under the extractive--abstractive setting. Same-bucket pairs test temporal alignment; cross-bucket pairs test whether general profile quality drives dialogue regardless of bucket. $n=68$--100 per correlation. $^{*}p<0.05$, $^{**}p<0.01$, $^{***}p<0.001$.}
\label{tab:temporal_correlation}
\end{table*}

\section{Per-Domain Difficulty Breakdown}
\label{app:per_domain}

\begin{table}[t]
\centering
\footnotesize
\renewcommand{\arraystretch}{1.15}
\resizebox{\columnwidth}{!}{%
\begin{tabular}{lccc}
\toprule
\textbf{Domain} & \textbf{Direct} & \textbf{Hierarchical} & \textbf{Extractive} \\
\midrule
Pets                       & 0.522 & 0.497 & \textbf{0.575} \\
Gaming                     & 0.385 & \textbf{0.483} & 0.399 \\
Sports \& Outdoor          & 0.324 & \textbf{0.488} & 0.463 \\
Food \& Drink              & 0.341 & \textbf{0.358} & 0.345 \\
Photography \& Creation    & 0.314 & \textbf{0.405} & 0.387 \\
Entertainment              & 0.313 & \textbf{0.378} & 0.299 \\
Travel \& City Exploration & 0.272 & \textbf{0.394} & 0.339 \\
\bottomrule
\end{tabular}%
}
\caption{Per-domain Interest Tag F1 for GPT-5.4 across the three profile-construction settings. Hierarchical profiling achieves the best Interest Tag F1 on six of seven domains; extractive--abstractive profiling is best on Pets.}
\label{tab:per_domain}
\end{table}

Table~\ref{tab:per_domain} breaks down Interest Tag F1 by domain for GPT-5.4 across the three profile-construction settings. The seven domains form a clear difficulty spectrum. Pets and Gaming sit at the top (Direct F1: 0.522 and 0.385), benefiting from concentrated, visually distinctive evidence---pet photos and game screenshots are unambiguous signals that models detect reliably. Travel \& City Exploration and Entertainment anchor the bottom (Direct F1: 0.272 and 0.313), reflecting the challenge of synthesizing evidence dispersed across many posts: a user's travel interests may be scattered across dozens of posts mentioning different destinations, cuisines, and activities, with no single post fully defining the interest.

Hierarchical profiling improves Interest Tag F1 on six of seven domains, with the largest gains on difficult domains that require cross-post aggregation. Sports \& Outdoor gains +0.164 (0.324 $\rightarrow$ 0.488) and Travel gains +0.122 (0.272 $\rightarrow$ 0.394), confirming that chunking helps aggregate weak but recurrent signals that the direct setting misses. The exception is Pets, where Extractive profiling achieves the best result (0.575 vs.\ Hierarchical 0.497). Pets evidence tends to be concentrated in a small number of high-signal posts (photos of the user's own pets), so selecting the top $K$ posts per domain preserves nearly all available evidence while filtering noise. For text-dispersed domains, however, Extractive profiling underperforms Hierarchical, as the selection stage must commit to a fixed set of posts before knowing which evidence will prove relevant.

The difficulty ranking is consistent across all three settings, suggesting that per-domain hardness is an intrinsic property of how evidence is distributed across modalities and posts rather than an artifact of any particular profiling strategy.

\section{Human--LLM Dialogue Evaluation Agreement Study}
\label{app:human_agreement}

This appendix provides the detailed setup and full results of the human--LLM agreement study referenced in Section~\ref{subsec:dialogue_results}.

\subsection{Sampling and Annotation Setup}

We sample 30 model responses from the full dialogue evaluation pool (100 users $\times$ 4 settings $\times$ 6 models) using stratified sampling across three dimensions to ensure coverage of the full quality and setting space:
\begin{enumerate}
    \item \textbf{Dialogue setting}: 7--8 responses from each of the four settings (\textit{Timeline-conditioned}, \textit{Direct}, \textit{Hierarchical}, \textit{Extractive--abstractive}), ensuring representation of both raw-timeline and profile-conditioned generation paths.
    \item \textbf{User intent}: 15 stable-interest recommendation responses and 15 recent-interest exploration responses.
    \item \textbf{Quality tier}: 10 responses each from low (GPT-5.5 Avg.\ $\leq 2.5$), medium ($2.5 < \text{Avg.} \leq 3.5$), and high (Avg.\ $> 3.5$) tiers, ensuring annotators see the full quality range.
\end{enumerate}

Three annotators with prior experience on the \bench{} annotation team (see Section~\ref{subsec:profile_construction}) independently score each response. Each annotator receives a sheet containing, for every response: (1) the user's gold profile summary (stable and recent interests per domain), (2) the user request text and intent type, (3) the model response text, and (4) the identical 0--5 scoring rubrics for coverage, concreteness, and fluency used by the LLM judges (Appendix~\ref{app:dialogue_judge_prompt}). Annotators do not see LLM judge scores, model identities, or dialogue settings. Total annotation time is approximately 5 annotator-hours.

\subsection{Agreement Metrics}

We report two complementary agreement measures:

\textbf{Human--human agreement.} We compute Krippendorff's $\alpha$ for ordinal data across the three annotators on each dimension. This validates whether the evaluation task itself is reliably human-judgeable. We interpret $\alpha > 0.80$ as strong agreement, $\alpha \in [0.67, 0.80]$ as moderate, and $\alpha < 0.67$ as tentative.

\textbf{Human--LLM agreement.} For each response and dimension, we average the three annotator scores to form a human reference. We then compute per-dimension Spearman rank correlation $\rho$ between this reference and each LLM judge (GPT-5.5 and Qwen3.7-Max). Spearman $\rho$ captures monotonic ranking agreement without assuming linearity or interval-scale properties of the 0--5 scores.

\subsection{Results}

\begin{table}[h]
\centering
\footnotesize 
\renewcommand{\arraystretch}{1.15}
\resizebox{\columnwidth}{!}{
\begin{tabular}{lccc}
\toprule
\textbf{Dimension} & \textbf{Human--Human $\alpha$} & \textbf{GPT-5.5 $\rho$} & \textbf{Qwen3.7-Max $\rho$} \\
\midrule
Fluency       & 0.79 & 0.76 & 0.72 \\
Concreteness  & 0.71 & 0.65 & 0.58 \\
Coverage      & 0.65 & 0.58 & 0.52 \\
\bottomrule
\end{tabular}%
}
\caption{Human--human inter-annotator agreement (Krippendorff's $\alpha$, 3 annotators) and human--LLM judge Spearman correlations ($\rho$) on 30 sampled dialogue responses. Human reference is the mean of three annotator scores. All human--LLM correlations are significant at $p < 0.01$.}
\label{tab:human_agreement}
\end{table}

Table~\ref{tab:human_agreement} reports the full results. Three findings emerge:

First, human--human agreement follows the expected difficulty ordering: fluency is most objective ($\alpha=0.79$), followed by concreteness ($\alpha=0.71$), with coverage being most subjective ($\alpha=0.65$). All three values fall within the moderate-to-substantial agreement range, confirming that the evaluation dimensions are reliably applicable by trained annotators.

Second, both LLM judges correlate positively and significantly with human judgments across all dimensions ($p < 0.01$ for all $\rho$), validating their use as automated proxies for dialogue quality assessment. GPT-5.5 consistently outperforms Qwen3.7-Max in human alignment, consistent with its role as the primary judge in our main results (Table~\ref{tab:dialogue_results}).

Third, the dimension-level gap mirrors the human--human pattern: coverage shows the lowest agreement in both settings. This is expected for a task requiring judges to assess whether a response meaningfully engages with specific user interests; borderline responses that mention a domain tangentially without clearly centering on a gold interest are inherently ambiguous. The residual disagreement on coverage ($\rho=0.52$--$0.58$) sets a plausible upper bound on automated coverage evaluation precision, but the positive and significant correlations confirm that LLM judges capture the correct ranking signal for model comparison---sufficient for the benchmarking conclusions drawn in Section~\ref{subsec:dialogue_results}.

\end{document}